\pdfoutput=1

\documentclass[11pt]{article}

\usepackage[final]{acl}

\usepackage{times}
\usepackage{latexsym}

\usepackage[T1]{fontenc}

\usepackage[utf8]{inputenc}

\usepackage{microtype}

\usepackage{inconsolata}

\usepackage{graphicx}
\usepackage{natbib} 
\usepackage{subcaption}
\usepackage{algorithm}
\usepackage{algorithmic}
\usepackage{amsmath}
\usepackage{listings}
\usepackage{amsfonts}
\usepackage{amssymb}
\usepackage{makecell}
\usepackage{multirow}
\usepackage{enumitem}
\usepackage{amsthm}
\usepackage{bm}
\usepackage{subcaption}
\usepackage{tcolorbox}
\usepackage{booktabs}
\usepackage{colortbl}
\usepackage{pifont}

\title{AutoMedEval: Harnessing Language Models for \\Automatic Medical Capability Evaluation}

\author{
 \textbf{Xiechi Zhang\textsuperscript{1}},
 \textbf{Zetian Ouyang\textsuperscript{1}},
 \textbf{Linlin Wang\textsuperscript{1}\thanks{Corresponding author.}},
 \textbf{Gerard de Melo\textsuperscript{2,5}},
\\
 \textbf{Zhu Cao\textsuperscript{3}},
 \textbf{Xiaoling Wang\textsuperscript{1}},
 \textbf{Ya Zhang\textsuperscript{4,5}},
 \textbf{Yanfeng Wang\textsuperscript{4,5}},
 \textbf{Liang He\textsuperscript{1}}
\\
\\
 \textsuperscript{1}East China Normal University, \textsuperscript{2}Hasso Plattner Institute/University of Potsdam\\
 \textsuperscript{3}Tongji University, \textsuperscript{4}Shanghai Jiao Tong University, \textsuperscript{5}Shanghai AI Laboratory\\
 \small{
   \textbf{Correspondence:} \href{mailto:llwang@cs.ecnu.edu.cn}{llwang@cs.ecnu.edu.cn}
 }
}

\begin{document}
\maketitle
\begin{abstract}
With the proliferation of large language models (LLMs) in the medical domain, there is increasing demand for improved evaluation techniques to assess their capabilities. 
However, traditional metrics like F1 and ROUGE, which rely on token overlaps to measure quality, significantly overlook the importance of medical terminology.
While human evaluation tends to be more reliable, it can be very costly and may as well suffer from inaccuracies due to limits in human expertise and motivation. Although there are some evaluation methods based on LLMs, their usability in the medical field is limited due to their proprietary nature or lack of expertise.
To tackle these challenges, we present AutoMedEval, an open-sourced automatic evaluation model with 13B parameters specifically engineered to measure the question-answering proficiency of medical LLMs. The overarching objective of AutoMedEval is to assess the quality of responses produced by diverse models, aspiring to significantly reduce the dependence on human evaluation. Specifically, we propose a hierarchical training method involving curriculum instruction tuning and an iterative knowledge introspection mechanism, enabling AutoMedEval to acquire professional medical assessment capabilities with limited instructional data. Human evaluations indicate that AutoMedEval surpasses other baselines in terms of correlation with human judgments.

\end{abstract}

\section{Introduction}
The emergence of increasingly powerful large language models (LLMs) has sparked significant advances across a range of real-world applications, including the medical field. The advent of numerous medical LLMs, e.g., the Med-PaLM series of models~\cite{singhal2023large,singhal2023towards,tu2023towards}, MedAlpaca~\cite{han2023medalpaca}, MedLLaMA~\cite{wu2023pmc}, and DoctorGLM~\cite{xiong2023doctorglm}, highlights the need for reliable and comprehensive evaluation systems to assess and compare their performance. To date, comprehensively evaluating the capabilities of various medical LLMs remains a formidable challenge, due to the required medical expert knowledge and huge workload~\cite{singhal2023large,chang2023survey}.


\begin{figure}[t]
  \centering
\includegraphics[width=0.9\linewidth]{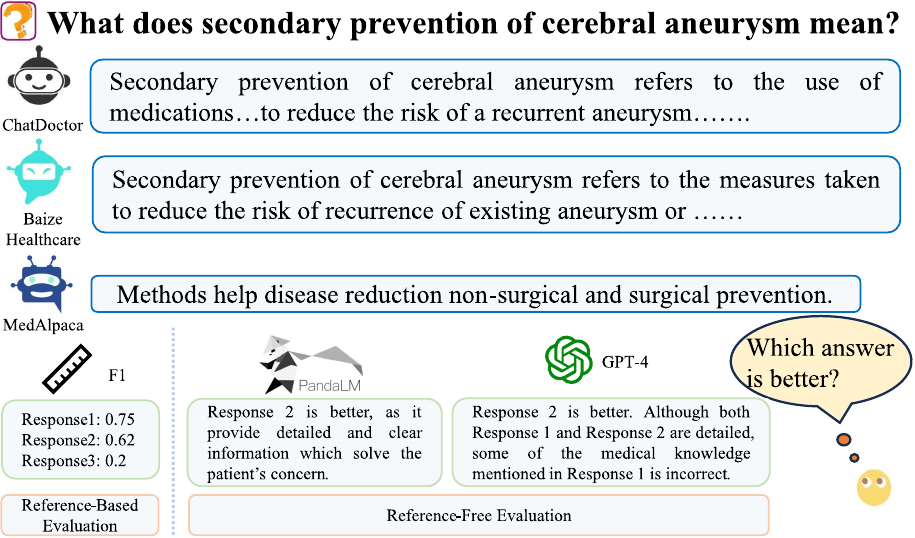}
\vspace{-1mm}
  \caption{Typical example of medical LLMs' responses evaluation.}
  \label{Figure 1}
  \vspace{-7.1mm}
\end{figure}

Human evaluation is often used to gauge the efficacy of medical LLMs, but this method is labor-intensive, time-consuming, and impractical on a large scale, especially when highly sought medical experts are needed~\cite{xu2023medgpteval}. As Figure~\ref{Figure 1} shows, traditional automatic evaluation methods, such as F1 and ROUGE~\cite{2004ROUGE}, tend to focus predominantly on lexical matches, frequently overlooking semantic nuances. While metrics based on pre-trained language models like BERTScore can assess at the sentence level, they do not align well with human judgment~\cite{liu-etal-2023-g, nie2024moca}. Benchmarks like USMLE~\cite{jin2021disease} can be used to assess the capabilities of medical LLMs, but they cannot perform open-ended generation evaluations under common clinical settings. Although proprietary models like GPT-4~\cite{nori2023capabilities} can provide detailed assessments, they are hampered by a lack of transparency and potential information leakage. Furthermore, GPT-4's performance on certain medical tasks is notably lower than that of human doctors~\cite{lai2023evaluating}, suggesting a low correlation with human judgment.

Leveraging open-sourced LLMs as evaluators represents an innovative and promising approach, already implemented in general evaluation domains. Although open-sourced evaluation models, such as PandaLM-7B~\cite{wang2023pandalm} and Auto-J~\cite{li2023generative} have been proposed, these models are designed for general scenarios. They often fall short in the medical domain, which necessitates specialized professional knowledge, making human evaluation essential but impractical at scale. This highlights the urgent need for an automatic, open-sourced evaluation model equipped with specialized medical expertise. However, incorporating accurate medical knowledge into the model and enabling it to perform detailed, human-like evaluations is challenging, especially when lacking large-scale, high-quality training data.

We propose AutoMedEval, a comprehensive evaluation model designed to provide detailed, human-aligned assessments, specifically intended to assist medical model developers in comparing the performance of different medical models. Detailed construction steps are as follows: (1) We first construct evaluation instructions from an existing medical QA dataset, developed using dynamic knowledge completion chains and validated by physicians through a double-checking procedure to ensure relevance for model training. (2) Based on the curated instructions, we leverage a hierarchical training approach to develop a novel LLM-based evaluation model, which is built upon the MedLLaMA-13B\footnote{https://huggingface.co/chaoyi-wu/MedLLaMA\_13B} model. 

Hierarchical training approach includes curriculum instruction tuning and iterative knowledge introspection phases, allowing the model to align more closely with human evaluation, even in the absence of large-scale high-quality data. \ding{192} The curriculum instruction tuning phase comprises three stages designed to imbue AutoMedEval with essential medical knowledge, enabling a deeper understanding of the evaluation task. \ding{193} The iterative knowledge introspection phase empowers AutoMedEval to continuously refine its evaluation accuracy by integrating revision suggestions derived from collaborative feedback between the model and physicians, establishing it as a reliable tool for evaluating medical LLMs' performance.

To sum up, our key contributions are as follows:
\begin{itemize}[noitemsep,nolistsep]
\item 
By leveraging a medical vector database with dynamic knowledge completions chains, we meticulously curate a high-quality medical instruction dataset. This dataset, rigorously validated by experienced physicians, serves as a robust foundation for the training of automatic evaluation models in the medical field.
\item  We propose an automated evaluation model called AutoMedEval, which is trained using a hierarchical training method and can introduce detailed, human-correlated evaluation of medical models. 
\item  Human evaluation and double-blind experiments are performed, which prove the effectiveness of the AutoMedEval model and the hierarchical training method.
\end{itemize}

\section{Task Formulation}
This task consists of two sub-tasks: (i) instruction dataset construction and (ii) training of an automated evaluation model. Given a QA dataset $T$ which consists of QA pairs $(q_{i}, a_{i})$ such that for each question $q_i$, each medical model $j$ in a set $S$ is used to generate a response $r_{i,j}$, leading to a combined $(q_{i}, r_{i,1}, ..., r_{i,|S|}, a_{i})$ for evaluation. After evaluating using a retrieval augmented LLM (GPT-4 and ChatGPT), we obtain the evaluation of each instance, which consists of a rationale text $e_{i}$ and a score $s_{i}$. Then the evaluation content and inputs are combined as instruction datasets in the format of $(q_{i}, r_{i,1}, ..., r_{i,|S|}, a_{i}, e_{i}, s_{i})$. The evaluation model $M$ trained using the instruction dataset should be able to rank each response $r_{i,j}$ and determine which medical model in $S$ is the best-performing one.


\begin{figure*}[htbp]
  \centering
\includegraphics[width=0.9\textwidth]{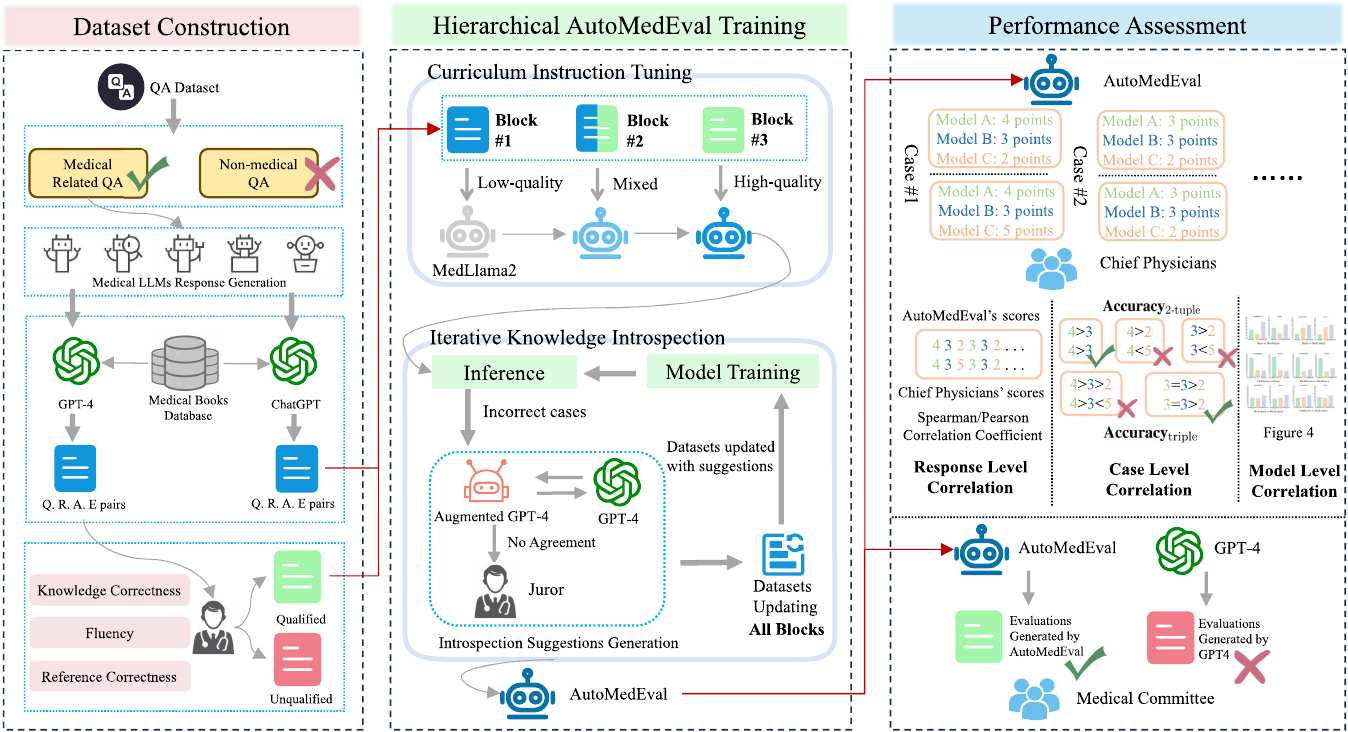}
  \caption{Creation of our instruction dataset and automatic evaluation model AutoMedEval.}
  \label{Figure 2}
  \vspace{-3mm}
\end{figure*} 

\section{Methodology}
\paragraph{Overview} As shown in the left part of Figure~\ref{Figure 2}, we first introduce the construction of a double checked instruction dataset based on vector databases and LLMs (Section~\ref{Instruction Data Collection}). Subsequently, as depicted in the middle part of Figure~\ref{Figure 2}, we present a hierarchical training strategy, including the Curriculum Instruction Tuning phase (Section~\ref{Curriculum Instruction Tuning}) and the Iterative Knowledge Introspection phase (Section~\ref{Iterative Knowledge Introspection}). It is designed to effectively integrate the medical knowledge and evaluative criteria from the instruction dataset into the model and calibrate the model's evaluation with human standards.

\subsection{Retrieval Augmented Instruction Dataset}
\label{Instruction Data Collection}

As illustrated in the left part of Figure~\ref{Figure 2}, to facilitate training automatic evaluation models in the medical field, we construct an instruction dataset based on 10K open-domain medical question answers pairs $T$ sourced from
medical-meadow-wikidoc\footnote{https://huggingface.co/datasets/medalpaca/medical\_me\-adow\_wikidoc} dataset. Specifically, we screen out a set $F_1$ of 126 cases with medically irrelevant content using pattern matching (example can be seen in Appendix Table~\ref{tab:data filter}), leaving 9,874 cases ($D = T \backslash F_1$) including topics shown in Appendix Figure~\ref{Figure Topics}. Then we exploit medical LLMs, including ChatDoctor~\cite{li2023chatdoctor} and Baize Healthcare~\cite{xu2023Baize}, to generate responses, and create ``\emph{patient's question}--\emph{responses}--\emph{answer}” $(q_{i}, r_{i,1}, r_{i,2}, a_{i})$ tuples as the inputs to GPT-4 for evaluation.

\begin{table}[t]
  \centering
    \begin{tabular}{c}
         \includegraphics[width=0.8\linewidth]{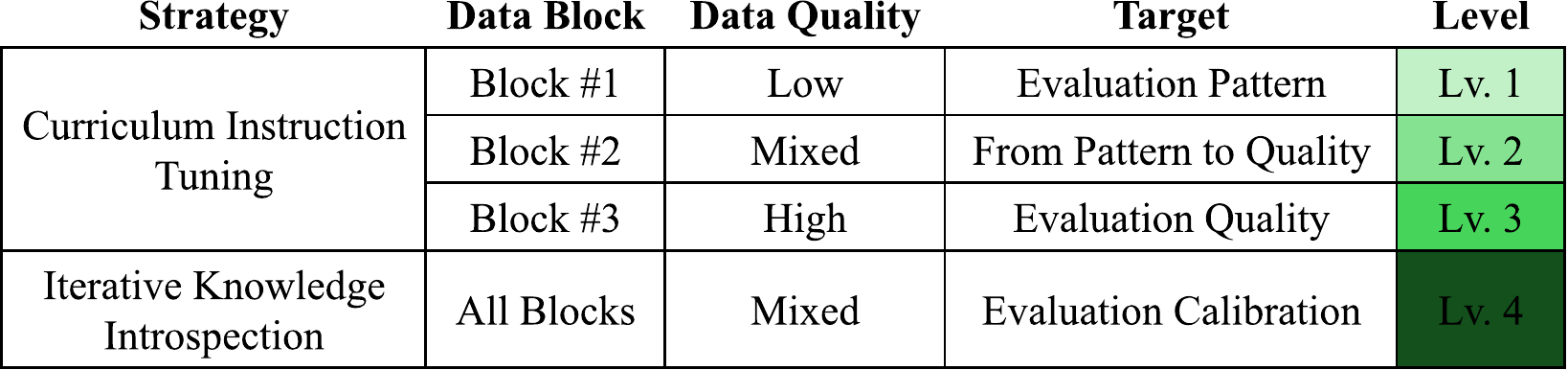}
    \end{tabular}
  
  \vspace{-2mm}
  \caption{Different stages of hierarchical training.}
  \label{FigureHierarchical}
  \vspace{-6mm}
\end{table}


To craft domain-specific instruction data, we adopt the common practice~\cite{wang2023pandalm} of distilling data from GPT-4 as evaluative evidence. 
We compile a collection of books and manuals shown in Appendix Table~\ref{detail of books and manuals}, encompassing medical knowledge and evaluation guidelines, to build a vector database. 
Subsequently, we utilize a customized prompt (Appendix Figure~\ref{Figurecom}) with one-shot prompting to optimize GPT-4's adaptation and propose a novel dynamic knowledge completion chain (Algorithm~\ref{alg:Knowledge Completion Chain}) to boost the reasoning ability of GPT-4.
Specifically, the instruction enables GPT-4 to generate a query starting with ``{\bf [Question]}” when uncertain about professional nouns or evaluation, and then our retrieval-augmented architecture will vectorize the query to provision corresponding knowledge from the vector database.
The extracted evidence 
aids in forming a dynamic completion 
chain that guides GPT-4 to deliver more credible high-quality medical evaluations. 


\begin{algorithm}[t]
\small
\caption{Knowledge Completion Chain}\label{alg:Knowledge Completion Chain}
\begin{algorithmic}[1]
\STATE \textbf{Input:} \text{$\mathcal{D}$ = Selected medical-meadow-wikidoc data.} 
\STATE  \text{$\mathcal{T}$ = Maximum number of  questions.}
\STATE \textbf{Output:} \text{$E$ = Dataset with synthesized evaluations.}
    \FOR{$i \xleftarrow{} 1$ to $|\mathcal{D}|$}
        \STATE $d \xleftarrow{} \mathcal{D}_i$
        \FOR{$t \xleftarrow{} 1$ to $\mathcal{T}$}
            \STATE $e \xleftarrow{} \text{GPT}(d)$
            \IF{``Question'' in $e$}
                \STATE $k_{i,t} \xleftarrow{} \text{Query(DB$_\text{Pinecone}$, } e)$
                \STATE $d \xleftarrow{} (d , e , k_{i,t})$
            \ELSE
                \STATE $E \xleftarrow{} E \cup (\mathcal{D}_i, e)$
                \STATE \textbf{break}
            \ENDIF
        \ENDFOR
    \ENDFOR

\end{algorithmic}
\end{algorithm}

\paragraph{Reliablity Verification}After collecting the evaluations, we engage two chief physicians to scrutinize the evaluation content of each case about (1) the correctness of medical knowledge involved when analyzing each model’s response. (2) whether there are instances of misattribution when analyzing each model’s response. (3) the fluency of language and the ease of understanding during the analysis. Any instruction data failing to satisfy the above criteria are regarded as invalid and discarded. After verification, the proportion of samples that meet the standard for each sub-aspect of the evaluation content in set $E$ is 94.06\%, 90.71\%, and 94.04\%, corresponding to the previous criteria.
Ultimately, a set $F_2$ of 305 unqualified cases is excluded, leaving a set comprising 9,569 entries ($R = E \backslash F_2$) and an example of the instruction is shown in the Appendix Figure~\ref{Figure 4}.

\subsection{Curriculum Instruction Tuning}
\label{Curriculum Instruction Tuning}
To imitate the lack of high-quality tuning data, we draw instead on ChatGPT to generate 9,569 evaluations $S$ on the same data.  Then we use a classifier (details can be found in Appendix~\ref{detail of the calssifier}) to obtain 4,788 high-quality instructions $R^{'}$ from GPT-4 and 3,823 relatively high-quality instructions $S^{'}$ from ChatGPT, with no intersection between the two sets of data, to train the model. 

As depicted in the middle part of Figure~\ref{Figure 2}, we use MedLLaMA as the foundation model. Inspired by the concept of cognitive synergy~\cite{goertzel2017toward}, we first train the AutoMedEval model using the curriculum instruction tuning strategy.
As shown in the top part of Table~\ref{FigureHierarchical}, the curriculum learning strategy comprises three-stage instruction tuning which incorporates critical medical and evaluation knowledge into the MedLLaMA model. Specifically, we randomly select 1,911 evaluations $S_{1}^{'}$ from $S^{'}$ as the instructions for curriculum \texttt{\#1}, which aims to help the model recognize evaluation patterns. Then we choose 2,394 evaluations $R_{3}^{'}$ from $R^{'}$ as curriculum \texttt{\#3} for steering the model to focus on the evaluation quality. Finally, we combine the remaining evaluations ($R^{'} - R_{3}^{'} + S^{'} - S_{1}^{'}$) to form the instructions for curriculum \texttt{\#2}, which ensures the model's transformation from patterns to quality. Therefore, MedLLaMA is trained sequentially with \texttt{\#1}, \texttt{\#2}, and \texttt{\#3}, and 
the training objective is formulated for model parameters $\widehat{\theta}$: 
\begin{equation}
\label{eq1}
   \widehat{ \theta^{*}}=\arg\!\max _{\widehat{\theta}} \sum_{j=1}^{N} \log p\left(Y^j \mid X^j; I^j, \phi\right)
\end{equation}

\noindent where $I=\{I_{c};I_{m};I_{h}\}$ refers to curriculum-based instructions that belong to curriculum \texttt{\#1}, \texttt{\#2}, and \texttt{\#3}, respectively.
$X^j$ and $Y^j$ denote the inputs and generated evalution of 
the $j$-th training instance. 
$\phi$ denotes the remaining parameters of the model.

\begin{algorithm}[htbp]
\caption{Iterative Knowledge Introspection}\label{alg:Iterative Knowledge Refinement}
\begin{algorithmic}[1]
\small
\STATE \textbf{Input:} \text{$\mathcal{R}$=$R^{'} + S^{'}$, $\mathcal{M}$=Previous trained model.}
\STATE  \text{$\mathcal{T}$=Number of iterations.}
\STATE \textbf{Output:} \text{Trained AutoMedEval model $\mathcal{M}$.}
    \FOR{$t \xleftarrow{} 1$ to $\mathcal{T}$}
        \STATE $\mathcal{G} \xleftarrow{} \mathcal{M}(\mathcal{R})$
        \STATE $\mathcal{I} \xleftarrow{} \text{Evaluate}(\mathcal{G})$ // $\mathcal{I}$:\text{ incorrect cases}
        \FOR{$i \xleftarrow{} 1$ to $|\mathcal{I}|$}
            \STATE $s_i \xleftarrow{} \text{Agent-Human Conversation}(\mathcal{I}_i)$
            \STATE $\mathcal{I}_i \xleftarrow{} (\mathcal{I}_i , s_i)$
        \ENDFOR
        \STATE $\mathcal{R^{'}} \xleftarrow{} \text{Update}(\mathcal{R}, \mathcal{I})$
        \STATE $\mathcal{M} \xleftarrow{} \text{Train}(\mathcal{R^{'}}, \mathcal{M})$
    \ENDFOR
\end{algorithmic}
\end{algorithm}

\begin{figure}[htbp]
  \centering
\includegraphics[width=0.75\linewidth]{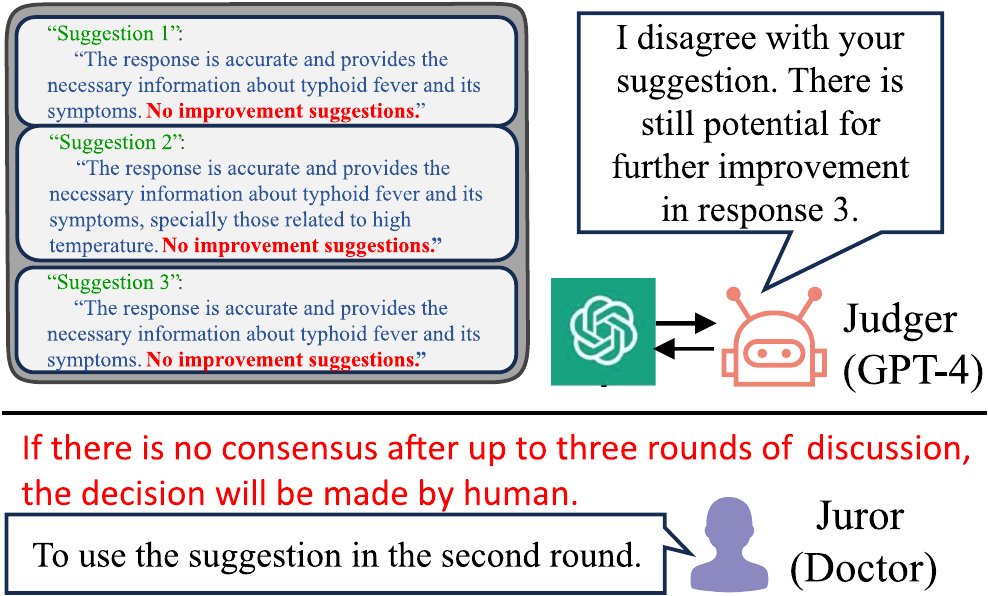}
  \caption{Collaborative Knowledge Introspection}
  \label{Figure Suggestion}
  \vspace{-6mm}
\end{figure}

\begin{table*}[!h]
\centering
\resizebox{0.9\linewidth}{!}{
\begin{tabular}{lcccccc}
\Xhline{1.0pt}

\multirow{2}[0]{*}{\textbf{Evaluation Models}} & \multicolumn{2}{c}{\textbf{Rationale Evaluation}} & \multicolumn{4}{c}{\textbf{Score Evaluation}} \\
\cmidrule(lr){2-3}\cmidrule(lr){4-7}
& \textbf{BERTScore} & \textbf{BARTScore} & \textbf{Spearman} & \textbf{Pearson} & \textbf{Accuracy$_{\rm 2\text{-}tuple}$} & \textbf{Accuracy$_{\rm triple}$} \\
\hline 
\rowcolor{gray!25}\multicolumn{7}{l}{\textit{Closed-sourced models}} \\
text-davinci-003 & 88.47$^{*}$ & -2.7589$^{*}$ & 0.3267$^{*}$ & 0.5459$^{*}$ & 47.92$^{*}$ & 17.11$^{*}$ \\ 
ChatGPT (gpt-3.5-turbo-0125) & 94.58$^{*}$ & -1.7301$^{\dagger}$ & 0.4856$^{*}$ & 0.5216$^{*}$ & 64.38$^{*}$ & 32.12$^{*}$ \\ 
GPT-4 (gpt-4-turbo-2024-04-09) & N/A & N/A & 0.5128$^{*}$ & 0.5689$^{*}$ & 67.98$^{*}$ & 35.42$^{*}$ \\ 
Gemini (gemini-2.0-flash) & 93.67$^{*}$ & -2.0143$^{*}$ & 0.5674$^{*}$ & 0.6082$^{*}$ & 71.68$^{*}$ & 42.63$^{*}$ \\ 
\hline
\rowcolor{gray!25}\multicolumn{7}{l}{\textit{Open-sourced models}} \\
Vicuna-7B & 86.94$^{*}$ & -2.7438$^{*}$ & 0.2669$^{*}$ & 0.2474$^{*}$ & 56.48$^{*}$ & 30.17$^{*}$ \\
Vicuna-13B & 87.86$^{*}$ & -2.7609$^{*}$ & 0.4766$^{*}$ & 0.4908$^{*}$ & 67.64$^{*}$ & 41.87$^{*}$ \\ 
DeepSeek-R1-Distill-Llama-8B & 94.02$^{*}$ & -1.9054$^{*}$ & 0.5527$^{*}$ & 0.6148$^{*}$ & 71.06$^{*}$ & 41.52$^{*}$ \\ 
MedLLaMA & 85.84$^{*}$ & -2.7528$^{*}$ & 0.3223$^{*}$ & 0.4848$^{*}$ & 55.18$^{*}$ & 26.23$^{*}$ \\ 
PandaLM	& N/A & N/A & 0.4311$^{*}$ & 0.4196$^{*}$ & 67.27$^{*}$ & 36.62$^{*}$\\
Ours (AutoMedEval) & \textbf{94.72} & -1.7136 & \textbf{0.6314}$^{*}$ & \textbf{0.6854}$^{*}$ & \textbf{74.61} & \textbf{48.65} \\
\hline
\rowcolor{gray!25}\multicolumn{7}{l}{\textit{Ablation Study}} \\
w/o Knowledge Completion Chain & 93.98$^{*}$ & -1.7139 & 0.4998$^{*}$ & 0.5510$^{*}$ & 63.76$^{*}$ & 27.38$^{*}$ \\
w/o Curriculum Instruction Tuning & 94.69$^{\dagger}$ & -1.7159 & 0.5375$^{*}$ & 0.6026$^{*}$ & 67.61$^{*}$ & 38.88$^{*}$ \\ 
w/o Iterative Knowledge Introspection & 94.62$^{*}$ & -1.6758$^{*}$ & 0.5864$^{*}$ & 0.6461$^{*}$ & 70.62$^{*}$ & 44.61$^{*}$ \\ 
\hline 
\rowcolor{gray!25}\multicolumn{7}{l}{\textit{Vanilla Instruction Tuning}} \\ 
5,000 GPT-4 instructions  w/ MedLLaMA & 94.13$^{*}$ & -1.7208$^{\S}$ & 0.5614$^{*}$ & 0.6251$^{*}$ & 66.02$^{*}$ & 31.88$^{*}$ \\ 
9,000 GPT-4 instructions  w/ MedLLaMA & 94.67$^{*}$ & \textbf{-1.6299}$^{*}$ & 0.5152$^{*}$ & 0.5832$^{*}$ & 67.96$^{*}$ & 35.13$^{*}$ \\ \hline
\end{tabular}
}
\caption{Comparison of results on different models. We run models three times and report the average results. * represents a significant difference with our results or significant correlation with human annotation (t-test, \emph{p}-value$<$ 0.001), while $\dagger$ and $\S$
refer to t-test with  \emph{p}$<$0.01 and \emph{p}$<$0.05, respectively. And the intraclass correlation coefficient (ICC) and Krippendorff’s alpha among five doctors is 0.712 and 0.725.}
\label{tab:main results}
\vspace{-5mm}
\end{table*} 

\subsection{Iterative Knowledge Introspection}
\label{Iterative Knowledge Introspection}
Subsequently, we introduce a technique shown in the Algorithm~\ref{alg:Iterative Knowledge Refinement} to mitigate AutoMedEval's incorrect evaluation. 
One potential reason for such inaccuracies could be that AutoMedEval 
harbors misconceptions about certain elements of evaluation knowledge. 
Inspired by the process of humans refining initial drafts with feedback~\cite{flower1981cognitive} and cognitive introspection
\cite{chen2023introspective,wang2023metacognitive}, we propose an iterative knowledge introspection approach to enhance the multi-step medical reasoning accuracy of our AutoMedEval model and calibrate it with human standards. This approach ultimately pushes the upper limit of its capabilities through AI-doctor collaboration. Specifically, we utilize our retrieval-augmented GPT-4 to generate revision suggestions and employ the standard GPT-4 as a judge to assess these suggestions.  As shown in Figure~\ref{Figure Suggestion}, if a consensus is not reached after up to three rounds of discussion, one chief physician will serve as the jury to make the final decision.
These suggestions are then used to update original instruction datasets and iteratively fine-tune our model for knowledge introspection.
Note that we generate revision suggestions exclusively for those training instances that were inaccurately predicted by the model from the previous iteration. The probability for the refined output $\widehat{Y}^j$ is formulated as 
\begin{equation}\label{eq:2}
\small
    \begin{aligned}
P{_\theta}=\sum_{Y^j} p\left(\widehat{Y}^j \mid X^j, Y^j,S; I^{j}, \phi\right) \,
    p\left(Y^j \mid X^j; I^{j}, \phi\right)
    \end{aligned}
\end{equation}
\noindent where $S$ refers to the revision suggestions from the AI-doctor collaborative feedback.

\section{Experiments}

\subsection{Baselines and Test Set}
Baselines for our experiments include models ranging from closed-sourced models (text-davinci-003, ChatGPT, GPT-4, Gemini) to open-sourced ones (PandaLM, Vicuna-7B/13B, Deepseek). The test set mainly includes 958 entries from Medical Meadow Wikidoc dataset and an additional 172 entries from MedText\footnote{https://huggingface.co/datasets/BI55/MedText} dataset. 
As for medical models evaluated, in addition to the ChatDoctor and Baize models mentioned earlier, MedAlpaca~\cite{han2023medalpaca} and MedLlama2\footnote{https://huggingface.co/llSourcell/medllama2\_7b}, which are not utilized in the initial response gathering phase, are employed to generate responses for the questions in the test set and then evaluated by AutoMedEval.


\subsection{Human Annotation and Judgement}
For evaluating AutoMedEval's capability, we engage five doctors to annotate the quality of responses generated by medical models in the aspects of \emph{Relevancy}, \emph{Fluency}, and \emph{Knowledge Correctness} and judge the evaluation generated by AutoMedEval in the aspects of \emph{Reference}, \emph{Fluency}, and \emph{Knowledge Correctness}. 
Then the average score of the metrics mentioned above is defined as the final score of each response or evaluation content. After annotation, we calculate the intraclass correlation coefficient (ICC) and Krippendorff’s alpha of five doctors to validate the annotation reliability. More annotation details can be found in Appendix \ref{detail of human annotation}.

\subsection{Evaluation Methods}

\paragraph{Score Evaluation}
As depicted in the right part of Figure~\ref{Figure 2}, we employ \textbf{Accuracy}$_{\rm 2\text{-}tuple}$ and \textbf{Accuracy}$_{\rm triple}$ to assess the correlation at the case level. \textbf{Accuracy}$_{\rm 2\text{-}tuple}$ measures the consistency between the relative magnitude of AutoMedEval's scoring and annotated scores while \textbf{Accuracy}$_{\rm triple}$ measures alignment over three medical LLMs. Besides, we use both \textbf{Spearman} and \textbf{Pearson} metrics to measure the correlation between AutoMedEval and humans at the response level. 

\paragraph{Rationale Evaluation}
We use two semantic evaluation metrics, \textbf{BERTScore}~\cite{zhang2019bertscore} and \textbf{BARTScore}~\cite{yuan2021bartscore}, to assist evaluation by using evaluations generated by GPT-4 as the reference answer.
 
\paragraph{Double-blind Preference}
A double-blind preference experiment is conducted to evaluate the practical applicability of AutoMedEval. Medical master candidates are instructed to make preference selections between the evaluation produced by AutoMedEval and the assessment results derived from GPT-4. More details can be seen in Appendix~\ref{detail of double blind preference experiment}.


\begin{figure*}[!h]
\centering
\small
\begin{subfigure}[b]{\linewidth}
    \centering
    \begin{subfigure}[b]{0.16\linewidth}
        \includegraphics[width=1\linewidth]{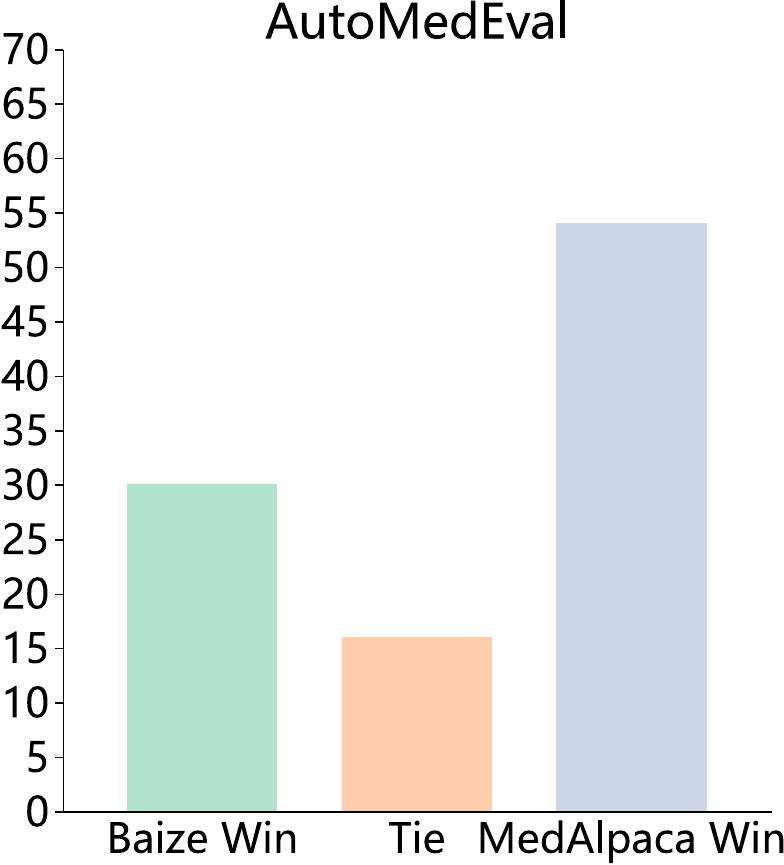}
    \end{subfigure}
    \begin{subfigure}[b]{0.16\linewidth}
        \includegraphics[width=1\linewidth]{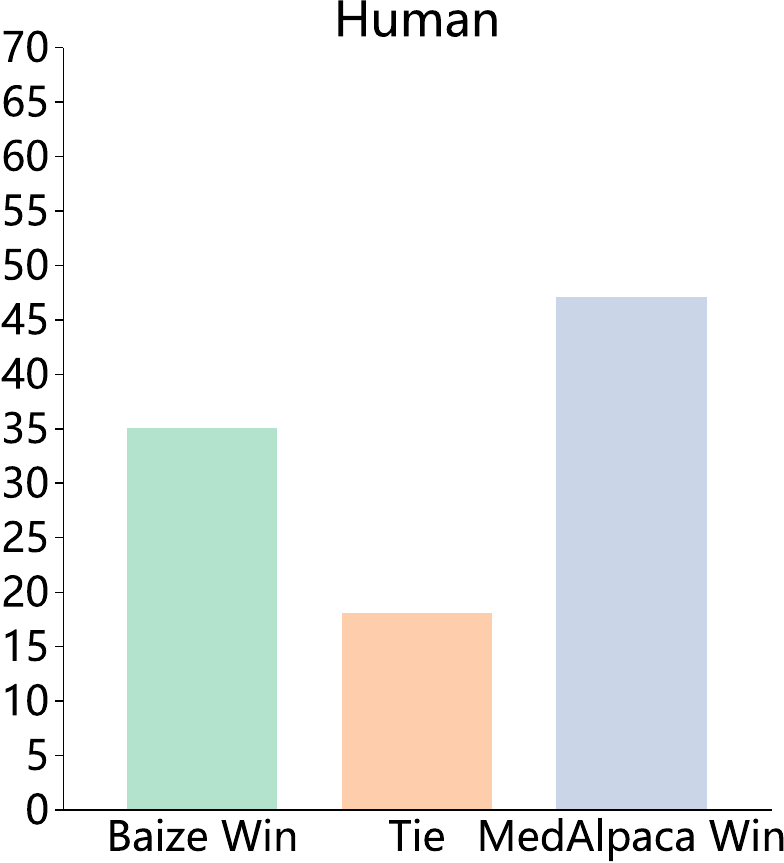}
    \end{subfigure}
    \hfill
    \begin{subfigure}[b]{0.16\linewidth}
        \includegraphics[width=1\linewidth]{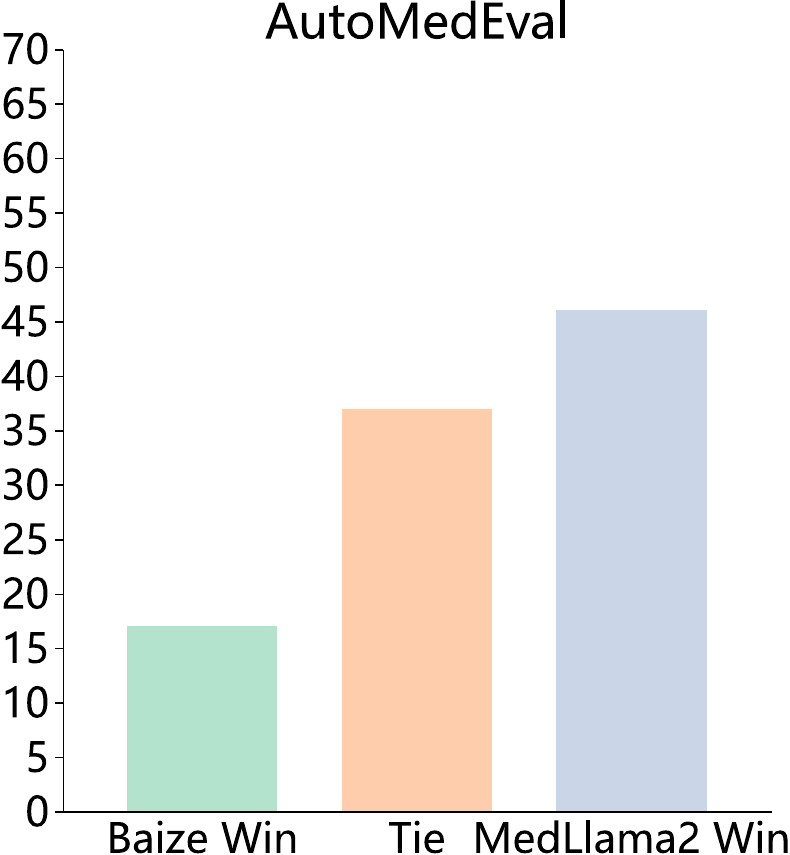}
    \end{subfigure}
    \begin{subfigure}[b]{0.16\linewidth}
        \includegraphics[width=1\linewidth]{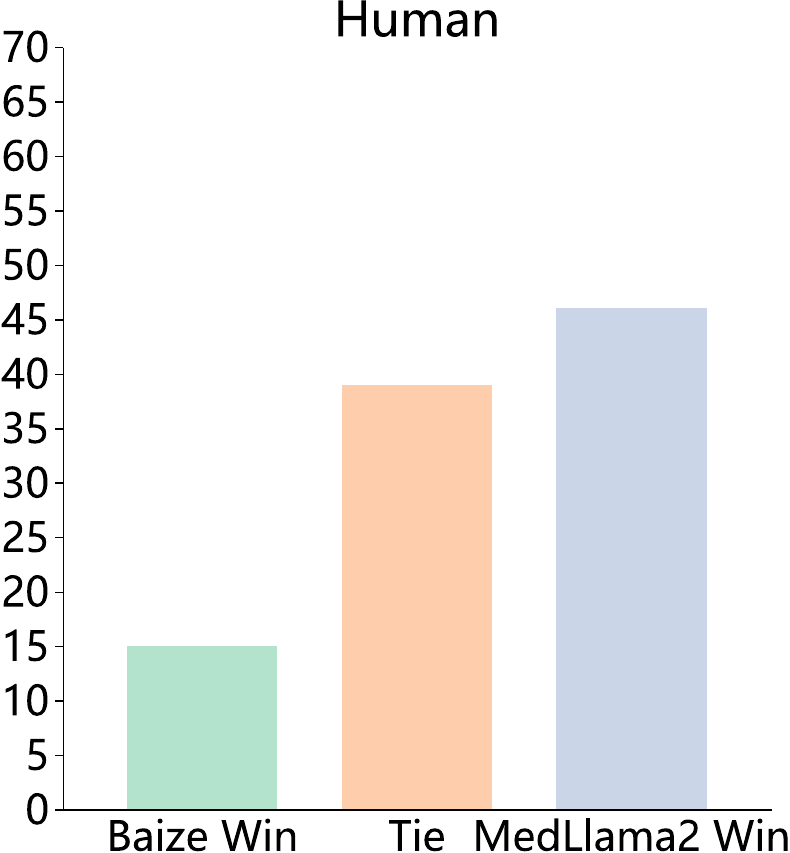}
    \end{subfigure}
    \hfill
    \begin{subfigure}[b]{0.16\linewidth}
        \includegraphics[width=1\linewidth]{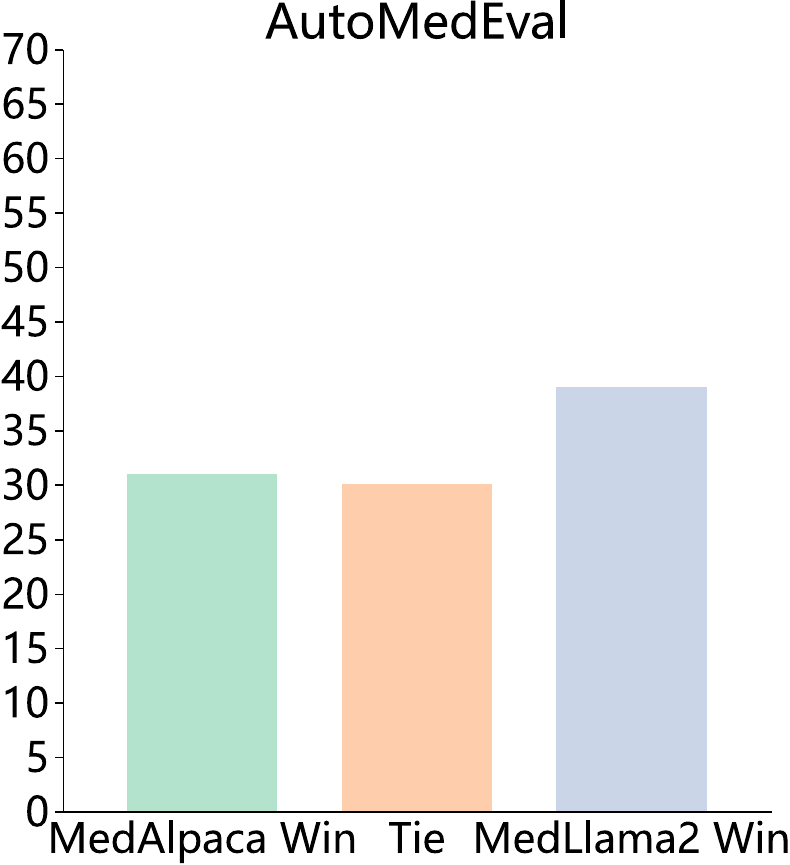}
    \end{subfigure}
    \begin{subfigure}[b]{0.16\linewidth}
        \includegraphics[width=1\linewidth]{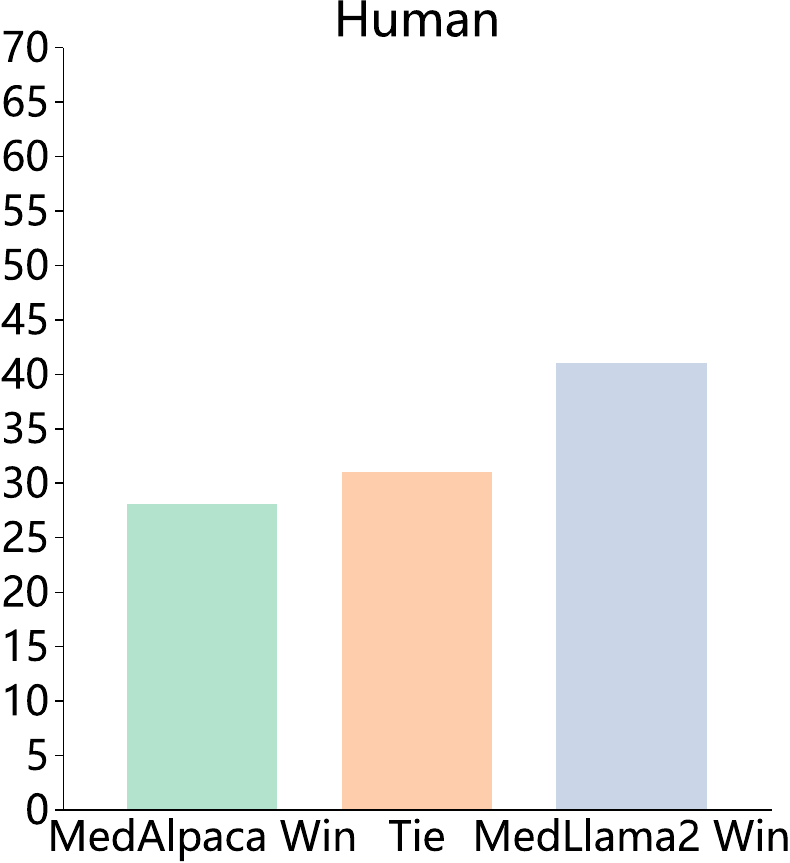}
    \end{subfigure}
\end{subfigure}

\vspace{0.5mm}

\begin{minipage}[b]{0.33\linewidth}
    \centering
    \text{Baize vs MedAlpaca}
\end{minipage}
\hfill
\begin{minipage}[b]{0.33\linewidth}
    \centering
    \text{Baize vs MedLlama2}
\end{minipage}
\hfill
\begin{minipage}[b]{0.33\linewidth}
    \centering
    \text{MedAlpaca vs MedLlama2}
\end{minipage}

\vspace{1.5mm}

\begin{subfigure}[b]{\linewidth}
    \centering
    \begin{subfigure}[b]{0.16\linewidth}
        \includegraphics[width=1\linewidth]{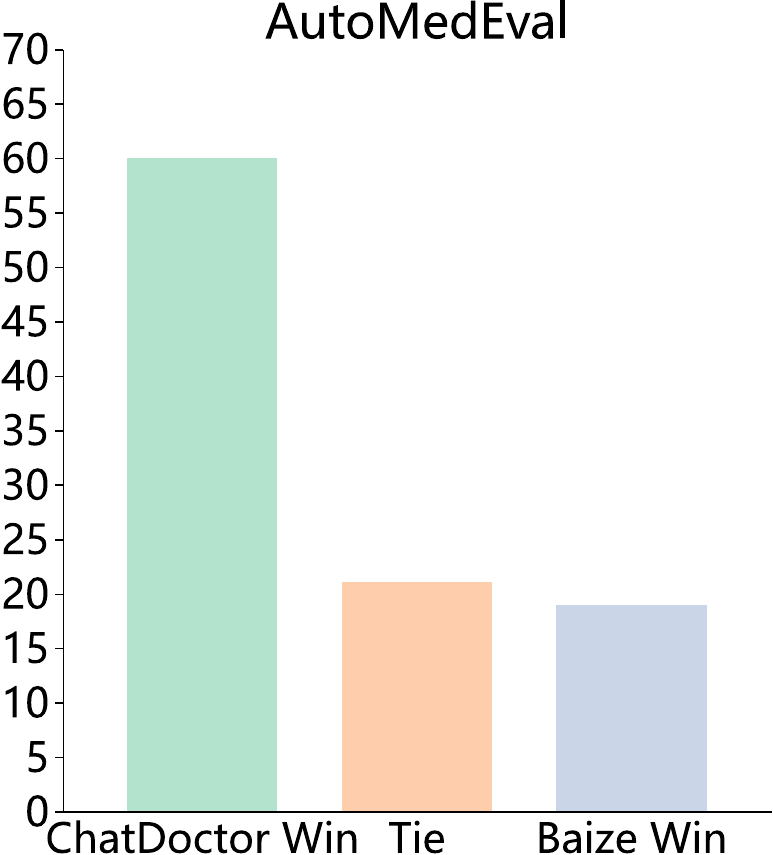}
    \end{subfigure}
    \begin{subfigure}[b]{0.16\linewidth}
        \includegraphics[width=1\linewidth]{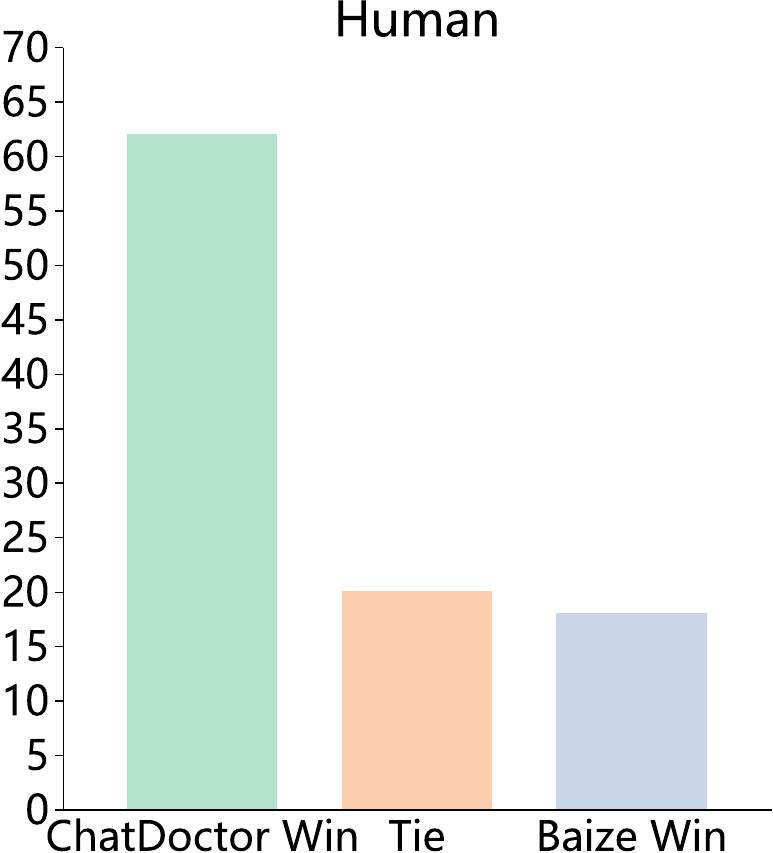}
    \end{subfigure}
    \hfill
    \begin{subfigure}[b]{0.16\linewidth}
        \includegraphics[width=1\linewidth]{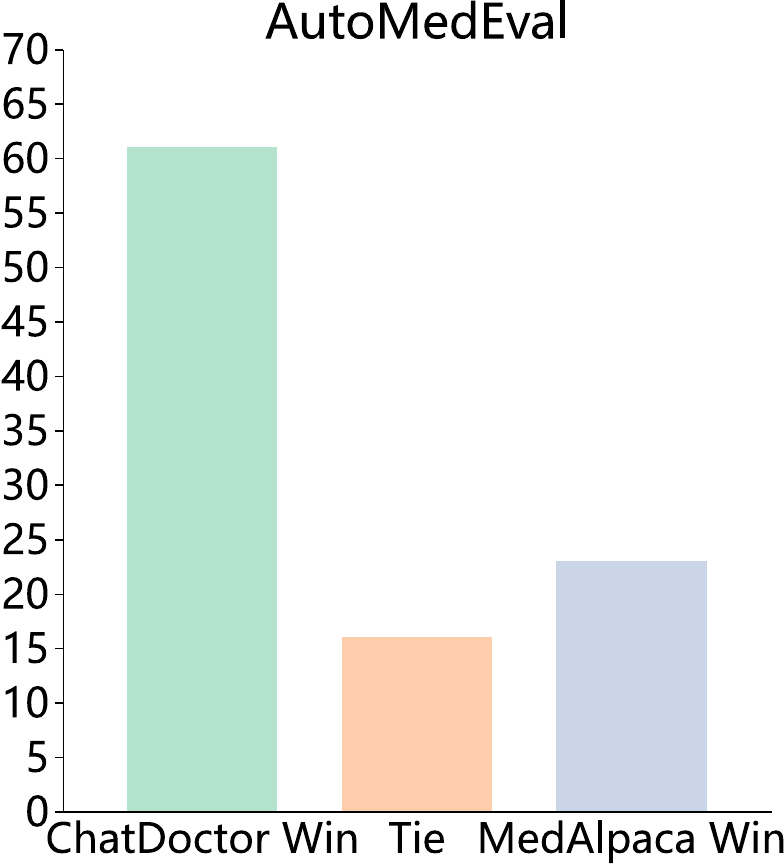}
    \end{subfigure}
    \begin{subfigure}[b]{0.16\linewidth}
        \includegraphics[width=1\linewidth]{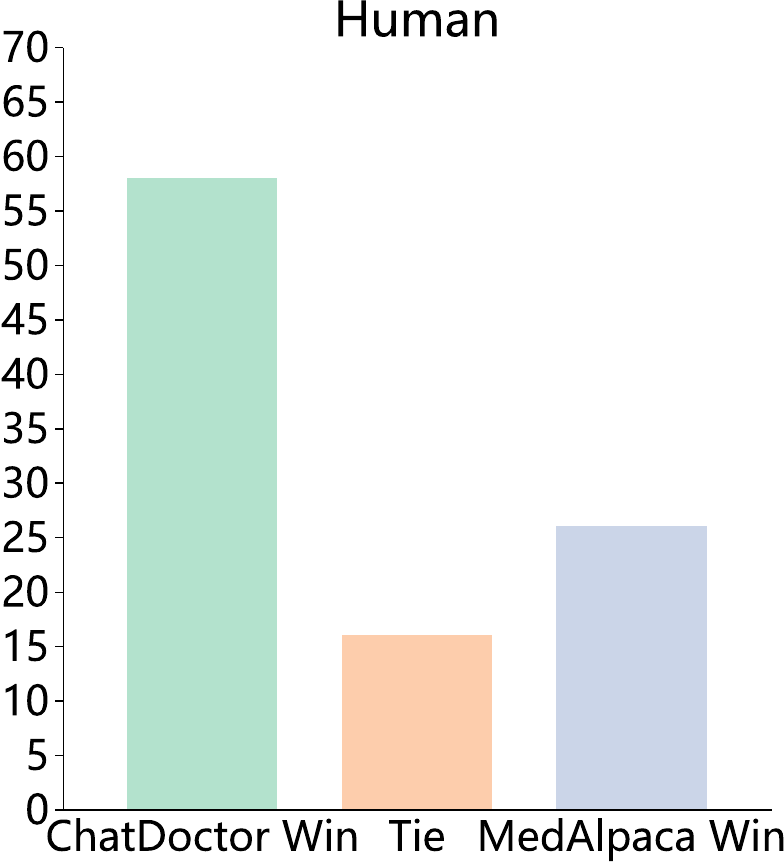}
    \end{subfigure}
    \hfill
    \begin{subfigure}[b]{0.16\linewidth}
        \includegraphics[width=1\linewidth]{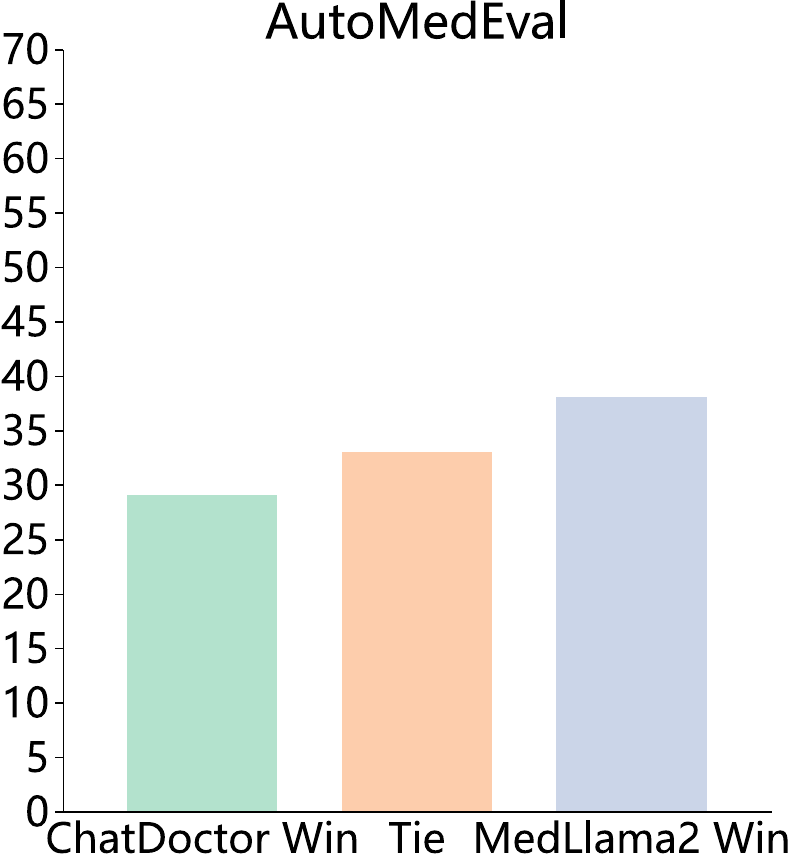}
    \end{subfigure}
    \begin{subfigure}[b]{0.16\linewidth}
        \includegraphics[width=1\linewidth]{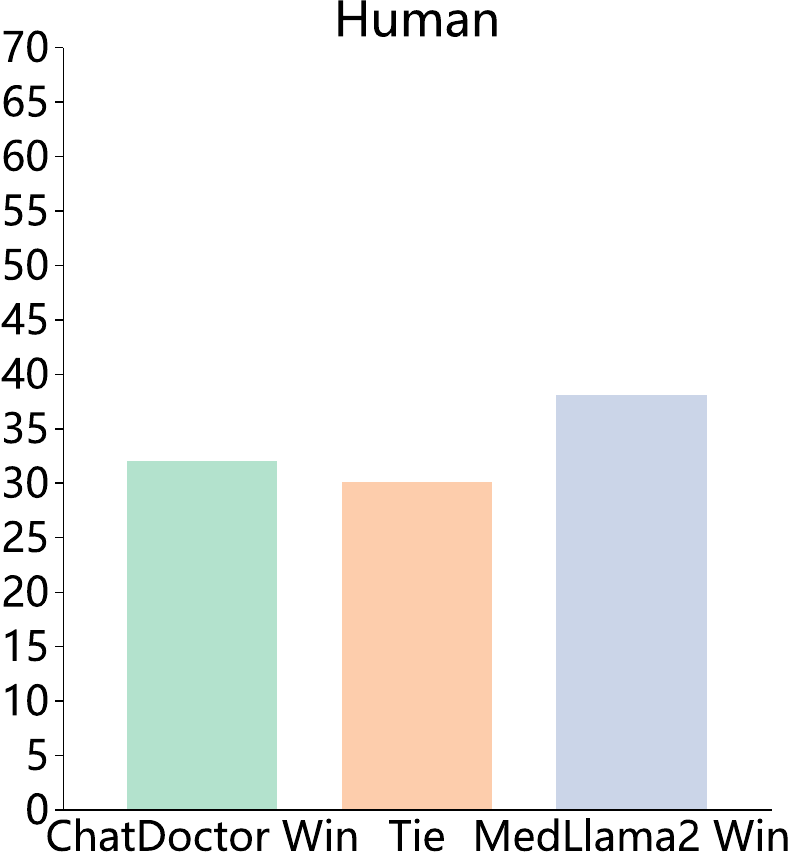}
    \end{subfigure}
\end{subfigure}

\vspace{0.5mm}

\begin{minipage}[b]{0.33\linewidth}
    \centering
    \text{ChatDoctor vs Baize}
\end{minipage}
\hfill
\begin{minipage}[b]{0.33\linewidth}
    \centering
    \text{ChatDoctor vs MedAlpaca}
\end{minipage}
\hfill
\begin{minipage}[b]{0.33\linewidth}
    \centering
    \text{ChatDoctor vs MedLlama2}
\end{minipage}
\vspace{-3mm}

\caption{AutoMedEval and chief physicians' judgment on every two medical LLMs' performance. "Win" means the ratio of cases where the current medical LLM outperforms another, while "Tie" means both medical LLMs' scores are the same.}
\label{compare result}
\vspace{-5mm}
\end{figure*}


\subsection{Main Results}
\label{main result}

\subsubsection{Score \& Rationale Evaluation Results}
To demonstrate the capability of AutoMedEval in evaluating medical models, we conducted a human-centered evaluation with the participation of chief physicians. The main results in Table~\ref{tab:main results} provide a comparison of AutoMedEval with representative baselines. 
We observe that AutoMedEval outperforms all other evaluation models, particularly surpassing GPT-4 by a significant margin across all metrics. Compared to PandaLM, an automatic evaluator in general domain, our AutoMedEval model achieves a relative improvement of 10.9\% on Accuracy$_{\rm 2\text{-}tuple}$. 

\subsubsection{Human-AutoMedEval Correlation}
We draw a comparative visualization in Figure \ref{compare result} that delineates the win rate comparisons between every pair of medical LLMs, as adjudicated by the chief physicians or the AutoMedEval model. A detailed examination of the win rate distributions within the evaluative outputs from AutoMedEval with those derived from human assessments reveals a notable congruence. This alignment suggests that AutoMedEval possesses the capability to discern the superior model with a degree of precision comparable to that of human evaluators, thereby underscoring the robust consistency between AutoMedEval and human annotators. Note that during the response generation phase, we use pairs of medical models to generate responses separately for each case. As shown in Appendix~\ref{GeneralSettings}, a sampling strategy is applied, causing the decoded response to vary for the same question under different cases, even for the same model. Therefore, performance transitivity does not apply to the results in Figure~\ref{compare result}.
Additionally, as illustrated in Figure~\ref{AutoMedEval evaluation and preference results} (a), the distribution of scores assigned by annotators to the evaluation content generated by AutoMedEval further demonstrates the experts' level of endorsement for it.


\subsubsection{Double-blind Experiment Results}

\begin{figure}[t]
\centering
\small
    \begin{subfigure}[b]{0.50\linewidth}
        \includegraphics[width=\linewidth]{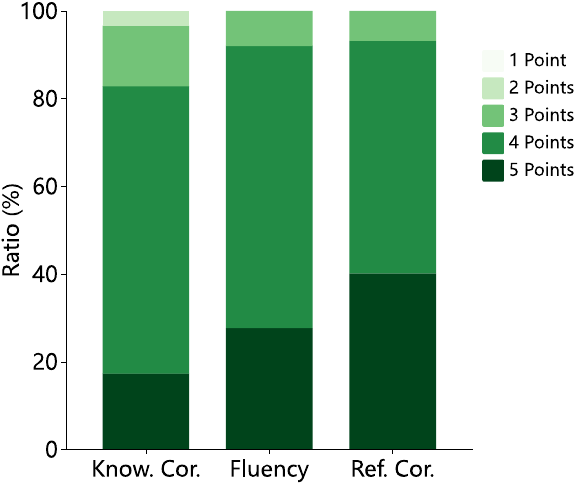}
    \end{subfigure}
    \begin{subfigure}[b]{0.46\linewidth}
        \includegraphics[width=\linewidth]{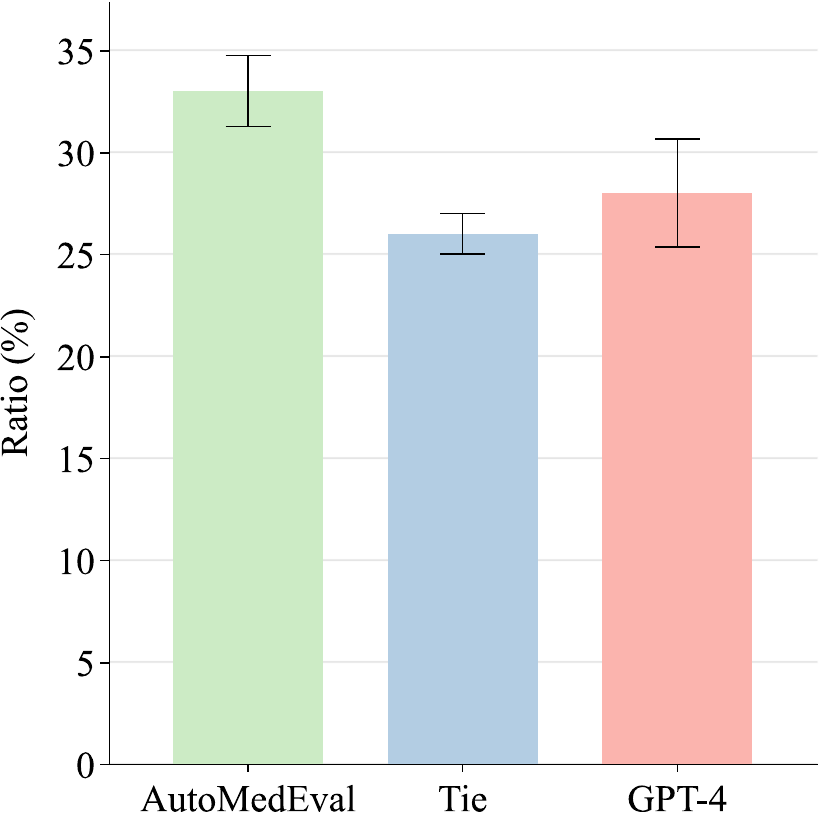}
    \end{subfigure}
    \begin{minipage}[b]{0.50\linewidth}
        \centering
        \text{(a)}
    \end{minipage}
    \hfill
    \begin{minipage}[b]{0.46\linewidth}
        \centering
        \text{(b)}
    \end{minipage}
    \caption{(a) Human assessment results on AutoMedEval's evaluation content. Know. Cor. represents Knowledge Correctness and Ref. Cor. means Reference Correctness. (b) Double-blind preference experiment results.}
    \label{AutoMedEval evaluation and preference results}
\vspace{-5mm}
\end{figure}

As illustrated in Figure \ref{AutoMedEval evaluation and preference results} (b), the evaluation outcomes generated by AutoMedEval garnered greater endorsement among medical professionals when contrasted with those produced by GPT-4. The error bars denote that despite the variance in preference selections among the three medical experts, the proportion of preferences accorded to AutoMedEval consistently surpassed those allocated to GPT-4.

\subsection{Quantitative Analysis}

\subsubsection{Ablation Studies}
To quantify the contributions of various components in AutoMedEval, we conducte ablation studies with three simplified architectures in Table~\ref{tab:main results}. We observe that all components contribute significant improvements. For example, the ablation of knowledge completion chain results in a relative 14.5\% degradation in Accuracy$_{\rm 2\text{-}tuple}$. Additionally, the removal of iterative knowledge introspection leads to a relative 5.3\% and 8.3\% decrease in Accuracy$_{\rm 2\text{-}tuple}$ and Accuracy$_{\rm triple}$ respectively.

\begin{table}[h]
    \centering
    \resizebox{0.7\linewidth}{!}{
    \begin{tabular}{c|ccc}
    \Xhline{1.0pt}
    \rowcolor{gray!25} {\textbf{Iteration Times}} & \textbf{0} & \textbf{1} & \textbf{2}\\ 
    \hline 
    \textbf{Accuracy$_{\rm triple}$} & 44.61 & 47.13 & 48.65 \\
    \textbf{Accuracy$_{\rm 2\text{-}tuple}$} & 70.80 & 73.01  &  74.61 \\
    \Xhline{1.0pt}
    \end{tabular}}
    \caption{Iteration results on the test set.}
    \vspace{-4mm}
    \label{fig:varying rounds and prompts}
\end{table}


\subsubsection{Effect of Varying Iterative Rounds}
To investigate the impact of varying iterative rounds in 
 knowledge introspection training, we conduct comparative experiments by training AutoMedEval with different iterations of knowledge introspection. As Table~\ref{fig:varying rounds and prompts} 
 depicts, an additional iteration of 
 knowledge introspection can enhance the performance of AutoMedEval.  For instance, after two iterations, our model outperforms curriculum instruction tuning by a relative 9.1\% improvement on Accuracy$_{\rm triple}$. 
We conduct mathematical modeling to determine when AutoMedEval achieves its best performance and find that AutoMedEval will outperform curriculum instruction tuning with a relative improvement of 17\% on the Accuracy$_{\rm triple}$ metric and there will not be any further growth 
after six iterations. Details of mathematical modeling are given in Appendix~\ref{mathematical modeling}. 


\subsubsection{Effectiveness under Limited Instructions}
The results in Table~\ref{tab:main results} also confirm the effectiveness of AutoMedEval 
using limited high-quality instructions. We observe that transitioning from 5,000 GPT-4 generated instructions to 9,000 instructions results in a significant improvement in Accuracy$_{\text{triple}}$ for MedLLaMA, indicating that increasing the number of high-quality instructions for tuning can enhance overall performance. Interestingly, AutoMedEval, utilizing 5,000 high-quality and 4,000 low-quality instructions, outperforms MedLLaMA with 9,000 high-quality instances by a relative improvement of 38.5\%. This demonstrates AutoMedEval's ability to address the scarcity of high-quality instructions by utilizing lower-quality data through our proposed instruction tuning.

\subsection{Qualitative Analysis}

\subsubsection{Case Study on Evaluations}
\begin{figure}[t]
    \centering
    \includegraphics[width=0.88\linewidth]{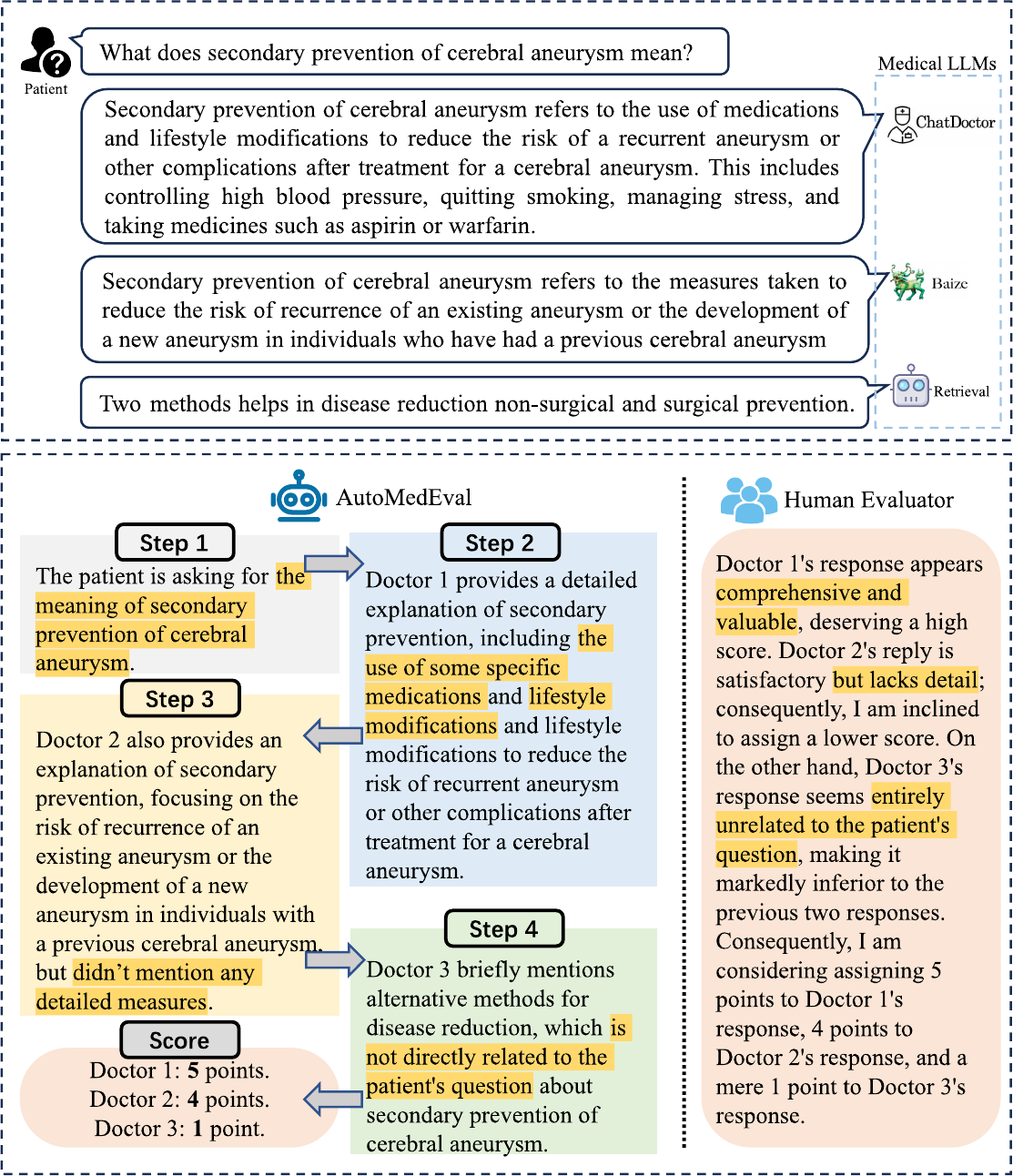}
    \caption{The rationale and score from AutoMedEval (left) and human annotator (right) for one example.}
    \label{Figure 7}
    \vspace{-4mm}
\end{figure}

As the case Figure~\ref{Figure 7} shows, during evaluation, our model first attempts to interpret the patient's question, and then evaluates each response according to the question. Eventually, it will assign scores to each response. In the given example, AutoMedEval first comprehends the patient's question and then evaluates each response about whether it explains the secondary prevention of cerebral aneurysm. This also entails comparing the content of different responses, such as the detailed measures mentioned in the first response, which are lacking in the second response. Thus, our model gives 5 points to ChatDoctor's response, 4 points to Baize Healthcare's response, and 1 point to Retrieval's response, which appears reasonable and accurate.

\subsubsection{Quality Inspection for Instructions}
For additional inspection of the data, we invoked LDA to model the topics in the remaining 9,569 instructions. We observe that the grouped topics are indeed highly task-related, including \emph{disease, medicine, diagnostics, treatment, prevention measures} and \emph{medical books}. 

Additionally, we conducted manual reviews of all non-consensus cases to check revision suggestions after receiving 
collaborative feedback. We observed that many instructions with incorrect medical knowledge were corrected, providing our knowledge introspection stage with more precise evidence. For the example in Figure~\ref{Figure Suggestion}, 
we obtained a new revision suggestion as \emph{``The response should include information about the symptoms of typhoid fever related to high temperature.”}

\section{Discussions}
\subsection{GPT-4 vs AutoMedEval}
With the development of LLMs, traditional metrics such as BLEU and ROUGE are no longer suitable for assessing the capabilities of LLMs in the NLG domain. Although proprietary models like GPT-4 have been proven to possess a certain level of general evaluation capability (\cite{wang2023chatgpt}, \cite{li2024leveraging}), ablation studies on GPT-4 reveal that its assessment ability in domains requiring specialized knowledge needs further enhancement. Moreover, due to their proprietary nature, models like GPT-4 are not suitable for use in professional fields with high privacy requirements, such as the medical field. In contrast, the AutoMedEval model we proposed, is an open-source evaluation model built upon a medical LLM backbone, which can effectively advance the development of evaluation models in the medical domain.


\subsection{Limitation of AutoMedEval}
We randomly sampled 20 cases to explore the limitations of AutoMedEval's medical knowledge. While it is evident that all evaluation content from AutoMedEval involves a certain level of medical knowledge, some inaccuracies can be categorized as follows: (1) incomplete expressions, (2) outdated information, (3) bias and hallucination,  (4) unsupported ratings. 

\section{Related Work}
Evaluation methodologies for medical LLMs generally fall into two categories: automatic and human evaluation. For automatic evaluation, fixed pre-defined metrics like precision, recall, and F1-score have long been used for information extraction~\cite{chinchor1993muc}.
The perplexity~\cite{manning1999foundations} metric was introduced to score 
the fluency of a model's output sentences. 
Comparisons with reference outputs can be obtained using BLEU~\cite{2002BLEU} and ROUGE~\cite{2004ROUGE}.
Most traditional automated metrics, however, assess the effectiveness of models solely at the lexical level, which is inadequate for more complex generation tasks, due to their failure to consider semantics and poor alignment with human judgments.

Human evaluations can be adjusted based on specific needs and expertise, but they may be prone to errors due to human limitations, especially in knowledge-intensive fields like medicine. Typically, human evaluation is used to support automatic evaluation results~\cite{belz2006comparing}. While human evaluation can be applied to entire datasets~\cite{xu2023medgpteval}, it is impractical at large scales due to the significant resources required, including time and money.

While LLMs continue to advance at a rapid pace, progress on automated evaluation methods to assess their performance has lagged behind. Although ChatGPT, GPT-4, and Gemini can aid automated assessments~\cite{nori2023capabilities, li2024leveraging, wang2023chatgpt}, they remain suboptimal options due to their proprietary nature and lack of reproducibility. Open-source evaluation models such as PandaLM and Auto-J are meant for generic tasks~\cite{wang2023pandalm} and obtain unsatisfactory results in the information-rich medical domain. Though some benchmarks have been proposed~\cite{pal2022medmcqa,jin2021disease}, it remains challenging to evaluate LLM's open-ended QA performance with benchmarks \cite{zheng2023judging}.

\section{Conclusion}
In this paper, we propose an instructions dataset and an automated evaluation model AutoMedEval that can facilitate the automatic evaluation of LLM in the medical field. Specifically, we use medical question-answer data and a novel dynamic knowledge completion chain method to collect evaluation results from the ChatGPT and GPT-4 models, which are then verified by chief physicians.

We further propose a hierarchical training strategy that including two techniques to train the model when lacking high-quality instruction data. One of these is curriculum learning, which uses lower-quality data and high-quality data to progressively bootstrap the model to gain evaluation capabilities. Another technique is iterative knowledge introspection, i.e., using training data to obtain extra suggestions, which helps the model align well with human judgment. Both training techniques benefit the model's performance in a series of comparative and ablation experiments.

Although AutoMedEval, trained on our new instruction dataset, demonstrates good performance, there is still some gap from practical applications. We will focus on collecting additional instruction data to develop superior evaluation models.


\section*{License}
The dataset, models, and tools used in this paper are open-sourced or permitted to be used for scientific research. The AutoMedEval model in the paper 
should only be used for research purposes.

\section*{Ethics Statement and Limitations}

We caution that the AutoMedEval model is primarily designed to help advance research and assess the performance of newly developed medical large language models in fundamental research settings. It is not intended to prove a medical model's suitability or effectiveness for genuine real-world deployment. As is the case for other automated text evaluation methods, the evaluations that our framework produces are merely shown to exhibit correlations with human judgments. However, despite a high correlation, such models can nevertheless make important mistakes. These obtained correlations are instead intended to help advance the state-of-the-art in research on medical language models. 

Another point that we wish to emphasize is the important role of data in medical language models. Special care was taken in this work to ensure that the data utilized is taken solely from open-source platforms that do not contain personal information.

\bibliography{custom}

\newpage
\appendix

\section{Example of Filtered Data}
\label{example of filtered data}
An example data item filtered through pattern matching string "rephras" is shown in Table~\ref{tab:data filter}.

\begin{table}[h]
    \centering
    \resizebox{0.7\linewidth}{!}{
    \begin{tabular}{c|c}
    \Xhline{1.0pt}
    \rowcolor{gray!25} \textbf{Composition} & \textbf{Content} \\
    \hline
    instruction & {\makecell[{{>{\raggedright}}p{9cm}}]{Answer this question truthfully}} \\
    \hline
    input   & {\makecell[{{>{\raggedright}}p{9cm}}]{Could you please provide me with the text that needs to be \textbf{rephras}ed? As "What is human DNA?" is already in proper English.}} \\
    \hline
    output & {\makecell[{{>{\raggedright}}p{9cm}}]{The prehistory period dates from around 7 x 10 6 b2k to about 7,000 b2k.}} \\
    \Xhline{1.0pt}
    \end{tabular}
    }
    \caption{Example data filtered through pattern matching.}
    \label{tab:data filter}
    \vspace{-5mm}
\end{table}

\section{Details of the Classifier}
\label{detail of the calssifier}
We use the verified GPT4 evaluations and the ChatGPT evaluations to train an evaluation quality classifier. Specifically, we initially label 200 cases with obvious reference or medical knowledge errors as ``low quality” and mark 200 verified evaluations generated by GPT-4 as ``high quality”. Subsequently, a contrastive learning framework SimCSE~\cite{gao2021simcse} is leveraged to derive the embeddings for each instruction. Ultimately, we use embedding-label pairs to train a support vector machine for classification and employ grid search to optimize the classifier's parameters. The quality classifier's accuracy on the 100 test cases is 91\% and is employed to distinguish the high-quality portions among the augmented instruction dataset.

\section{Experimental Settings}
\label{GeneralSettings}
We train our 13B AutoMedEval model using 8 NVIDIA A100 80GB GPUs and leverage DeepSpeed~\cite{rasley2020deepspeed} ZeRO-Stage 3~\cite{rajbhandari1910zero} for optimization. We use AdamW optimization with the WarmupDecay learning rate scheduler and set the learning rate to be 2$\times 10^{-5}$ with a warmup ratio of 0.03 for stability. Our model undergoes 5 training epochs with a batch size of 2 per GPU and a gradient accumulation step of 8. We adopt Flash attention~\cite{dao2022flashattention,dao2023flashattention} for efficient memory usage, thereby allowing a maximum input size of 2,048 tokens. For ChatDoctor, Baize Healthcare and other medical models during response generation, we set the temperature to 0.5 and the maximum number of new tokens as 200,  and apply a sampling strategy with top $k$ and top $p$ as 50 and 1, respectively.


\section{Annotation Details}
\label{detail of human annotation}
The expert team was composed of five doctors from different departments of tertiary hospitals consisting of three attending doctors and two chief physicians. An example of annoation results is shown in Table~\ref{tab: human annotation example}. For annotation, we developed and deployed a simple website that predefined all rating rules and requires the doctors to click on the relevant options. Before annotating, the following assessment standards were given to and read by each doctor: \\
\textit{Evaluation Metrics:
\begin{itemize}[noitemsep,nolistsep]
 \item  Relevancy assesses how well generated responses match the corresponding questions.
\item  Fluency evaluates naturalness and human-like quality of responses or AutoMedEval's
evaluations. 
 \item Knowledge Correctness is medical knowledge accuracy in responses or evaluations. 
 \item Reference Correctness assesses whether there are instances of misattribution in AutoMedEval's evaluation when quoting the medical LLMs.
\end{itemize}
Two tasks need to be completed:\\
(1) After each response, three scores need to be assigned, corresponding to the following three aspects:
\begin{itemize}
    \item Assess the relevance of each response to the question (whether it answers the question asked).
    \item Evaluate the correctness of medical knowledge contained in each response (appropriate use of terminology).
    \item Assess the fluency of language and the ease of understanding of each response.
\end{itemize}
(2) At the end of each sample, rate the following three aspects related to the "assessment":
\begin{itemize}
    \item Evaluate the correctness of medical domain knowledge applied when analyzing each doctor's response.
    \item Assess whether there are instances of misattribution when analyzing each doctor's response.
    \item Evaluate the fluency of language and the ease of understanding during the analysis.
\end{itemize}
Note: Please do not refer to the scores in the "assessment" section before completing the first part of the task.}

After annotation is completed, we perform an averaging operation on the scores assigned by all annotators for each annotation item to derive the final rating. 
Every annotator is paid for 20 CNY each case. 

\section{Detail of Double Blind Preference Experiment}
\label{detail of double blind preference experiment}
We invited three medical master candidates as the medical committee to participate in a double-blind preference experiment, with 87 randomly selected test samples. Before annotation, they were instructed to read the following guidelines:\\
\textit{After reading two evaluation results, you need to choose which one is better according the metrics listed below:
\begin{itemize}
    \item Knowledge Correctness: Evaluate the correctness of medical domain knowledge applied when analyzing each doctor's response.
    \item Reference Correctness: Assess whether there are instances of misattribution when analyzing each doctor's response (e.g., mistaking the content of the second doctor's response for the first doctor's response or unpractical information).
    \item Fluency: Evaluate the fluency of language and the ease of understanding during the analysis.
\end{itemize}
}

\section{Mathematical Modeling}
\label{mathematical modeling}
We utilize a variation of the sigmoid function to mimic the $P_t^{1,2,3}$ function as shown in Eq.~(\ref{eq:p1}), Eq.~(\ref{eq:p2}), and Eq.~(\ref{eq:p3}), which represents the probability each predicted score matches the ground truth.
\begin{equation}\label{eq:p1}
\small
    \begin{aligned}
    P^1_{t}=\frac{a_1}{1 + e^{-0.453t - 2.83}}
    \end{aligned}
\end{equation}
\begin{equation}\label{eq:p2}
\small
    \begin{aligned}
    P^2_{t}=\frac{a_2}{1 + e^{-0.453t - 2.83}}
    \end{aligned}
\end{equation}
\begin{equation}\label{eq:p3}
\small
    \begin{aligned}
    P^3_{t}=\frac{a_3}{1 + e^{-0.453t - 2.83}}
    \end{aligned}
\end{equation}

The accuracy of each sample is obtained by multiplying the three accuracy functions mentioned above, resulting in:
\begin{equation}\label{eq:ft}
    \begin{aligned}
    f(t)=\frac{a_1*a_2*a_3}{(1 + e^{-0.453t - 2.83})^3}.
    \end{aligned}
\end{equation}

\begin{table}[h]
    \centering
    \resizebox{0.5\linewidth}{!}{
    \begin{tabular}{c|c}
    \Xhline{1.0pt}
    \rowcolor{gray!25} {\textbf{Iteration Times}} & {\textbf{Accuracy$_{\rm triple}$}}\\ 
    \hline 
    Iter 0 & 44.61 \\
    Iter 1 & 47.13 \\
    Iter 2 & 48.65 \\
    \Xhline{1.0pt}
    \end{tabular}}
    \caption{Iteration results on the test set.}
    \vspace{-6mm}
    \label{tab: iteration result}
\end{table}
Using the data of iteration results in Table \ref{tab: iteration result}, we calculate the final accuracy estimation as follows.
\begin{equation}\label{eq:final}
\small
    \begin{aligned}
    f(t)=\frac{1*0.9*0.586}{(1 + e^{-0.453t - 2.83})^3}=\frac{0.5274}{(1 + e^{-0.453t - 2.83})^3}
    \end{aligned}
\end{equation}
When the iteration count is set to 6, the accuracy reaches 52.21\%.

\section{Instruction Examples}
Figures~\ref{Figurecom} and \ref{Figure 4} depict the instruction utilized to obtain evaluation data from GPT-4 and an example instruction exploited to tune the model, respectively.

\section{Example of Incorrect Evaluation}
An incorrect evaluation case with reference evaluation is depicted in Table \ref{tab:incorrect evaluation}.

\begin{figure}[h]
  \centering
\includegraphics[width=0.7\linewidth]{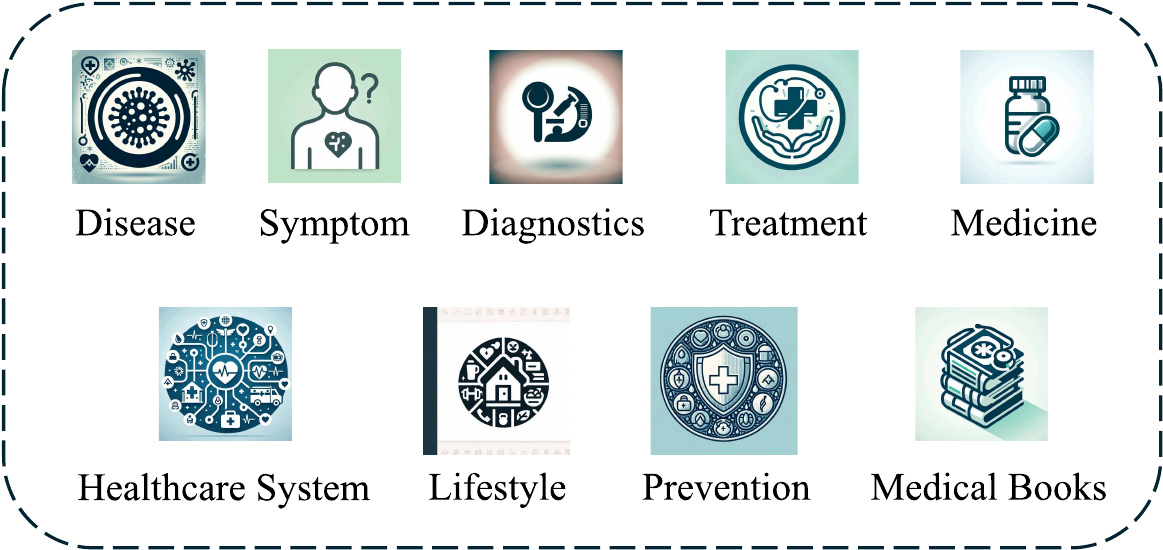}
  \caption{Topics covered by the dataset.}
  \label{Figure Topics}
  \vspace{-7mm}
\end{figure}

\begin{figure}[h]
  \centering
  \includegraphics[width=0.75\linewidth]{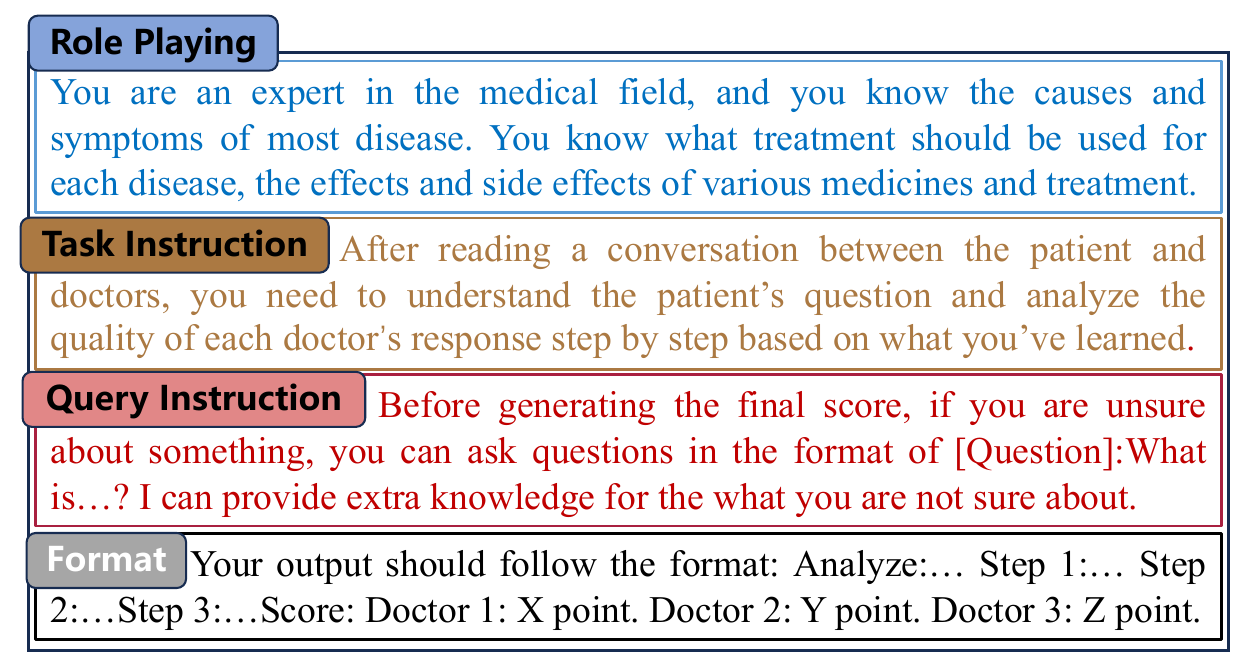}
   \caption{Instruction with diverse factors
  for GPT-4.}
  \label{Figurecom}
  \vspace{-7mm}
\end{figure}

\begin{figure}[H]
  \centering
  \includegraphics[width=0.8\linewidth]{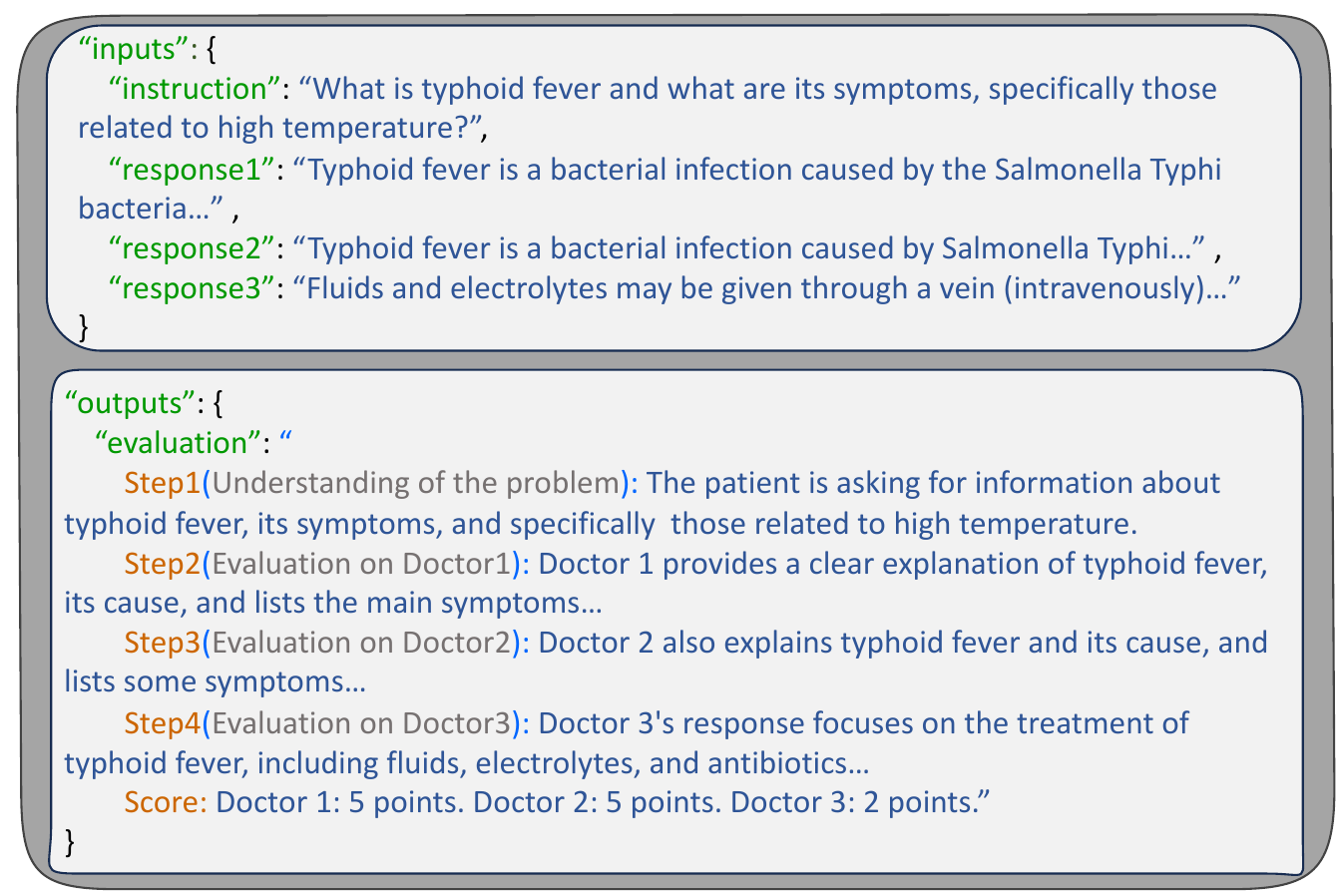}
  \caption{Example instruction for tuning AutoMedEval.}
  \label{Figure 4}
  \vspace{-8mm}
\end{figure}

\begin{table*}[htbp]
\small
    \centering
    \resizebox{0.98\textwidth}{!}{
    \begin{tabular}{l|r}
    \Xhline{1.0pt}
    \rowcolor{gray!25} \textbf{Medical Knowledge} & \textbf{Number of Pages}\\ 
    \hline 
    PFENNINGER \& FOWLOER'S Procedures for Primary Care & 1,763\\
    DECISION MAKING in Medicine & 753\\
    TEXTBOOK OF Physical Diagnosis HISTORY AND EXAMINATION & 965\\
    Primer of DIAGNOSTIC IMAGING & 808\\
    THORACIC IMAGING & 501\\
    SEIDEL'S GUIDE TO PHYSICAL EXAMINATION AN INTERPROFESSIONAL APPROACH & 676\\
    Differential Diagnosis OF Common Complaints & 703\\
    Diagnostic Imaging Interventional Procedures & 909\\
    DIAGNOSTIC ULTRASOUND & 2,249\\
    Roberts and Hedges' CLINICAL PROCEDURES in Emergency Medicine and Acute Care & 1,786\\
    Mosby's Manual of Diagnostic and Laboratory Tests & 1,201\\
    HENRY'S Clinical Diagnosis AND Management By Laboratory Methods & 1,956\\
    Grainger \& Allison's DIAGNOSTIC RADIOLOGY & 2,218\\
    Mosby's Diagnostic and Laboratory Test Reference & 1,095\\
    \hline 
    \rowcolor{gray!25} {\textbf{Evaluation Knowledge}} & \textbf{Number of Pages}\\
    \hline 
    Matrix specification of Core Clinical Conditions for the Physician Associate by category of level of competence & 127\\
    Professional Practice Evaluation Policy & 18\\
    SoP Consolidated Version  & 202\\
    \Xhline{1.0pt}
    \end{tabular}}
    \caption{Details of the books and manuals used to build the vector database.}
    \label{detail of books and manuals}
    \vspace{-5mm}
\end{table*}

\begin{table*}[htbp]
    \centering
    \resizebox{0.98\textwidth}{!}{
    \begin{tabular}{l|c}
    \Xhline{1.0pt}
    \rowcolor{gray!25} \textbf{Composition} & \textbf{Content} \\
    \hline
    {\makecell[{{>{\raggedright}}p{2cm}}]{Question}} & {\makecell[{{>{\raggedright}}p{20cm}}]{What information is available on Verapamil and Trandolapril?}} \\
    \hline
    {\makecell[{{>{\raggedright}}p{2cm}}]{Response 1\\(ChatDoctor)}} & {\makecell[{{>{\raggedright}}p{20cm}}]{Verapamil and Trandolapril are both medications used to treat high blood pressure, also known as hypertension. Verapamil is a calcium channel blocker that helps to relax the blood vessels and reduce blood pressure by preventing the constriction of the blood vessels. Trandolapril is an ACE inhibitor which works by blocking the enzyme responsible for converting angiotensin I into a more potent form, thus reducing blood pressure. Both medications can be taken alone or in combination with other medicines.  Common side effects of both drugs include dizziness, headache, fatigue, cough, and nausea. It is important to inform your doctor if you experience any serious side effects such as difficulty breathing, swelling, rash, or fever.}} \\
    \hline
    {\makecell[{{>{\raggedright}}p{2cm}}]{Annotator}} & {\makecell[{{>{\raggedright}}p{20cm}}]{Relevancy: 4 points. Knowledge Correctness: 4 points. Fluency: 4 points.}}\\
    \hline
    {\makecell[{{>{\raggedright}}p{2.3cm}}]{Response 2\\(Baize Healthcare)}} & {\makecell[{{>{\raggedright}}p{20cm}}]{Verapamil is a calcium channel blocker medication used to treat high blood pressure, chest pain, and certain types of irregular heartbeats. Trandolapril is an ACE inhibitor used to treat high blood pressure. Both medications have potential side effects and should be taken under the guidance of a healthcare provider.}} \\
    \hline
    {\makecell[{{>{\raggedright}}p{2cm}}]{Annotator}} & {\makecell[{{>{\raggedright}}p{20cm}}]{Relevancy: 3 points. Knowledge Correctness: 2 points. Fluency: 2 points.}}\\
    \hline
    {\makecell[{{>{\raggedright}}p{2cm}}]{Resposne 3\\(MedAlpaca)}} & {\makecell[{{>{\raggedright}}p{20cm}}]{The combination of verapamil and trandolapril is used to treat high blood pressure. It is a combination of two medications. It decreases certain chemicals that tighten the blood vessels, so blood flows more smoothly. It also relaxes your blood vessels so your heart does not have to pump as hard. This medication is sometimes prescribed for other uses; ask your doctor or pharmacist for more information.}} \\
    \hline
    {\makecell[{{>{\raggedright}}p{2cm}}]{Annotator}} & {\makecell[{{>{\raggedright}}p{20cm}}]{Relevancy: 2 points. Knowledge Correctness: 2 points. Fluency: 2 points.}}\\
    \hline
    {\makecell[{{>{\raggedright}}p{2cm}}]{Evaluation\\(AutoMedEval)}} & {\makecell[{{>{\raggedright}}p{20cm}}]{Analyze: \\
    Step 1: The patient is asking for information on Verapamil and Trandolapril.\\
    Step 2: Doctor 1 provides a comprehensive response, explaining the drug of both medications, their mechanism of action, and common side effects. The response also mentions the importance of informing a doctor if oftencing serious side effects.\\
    Step 3: Doctor 2 also provides information on the uses of Verapamil and Trandolapril, but the response is less detailed than Doctor 1's. The side effects mentioned are related to the medications, and the response emphasizes the importance of prescriptions under the care of a healthcare provider.\\
    Step 4: Doctor 3 explains the purpose of the combination between Verapamil and Trandolapril, but the response is cut off and does not provide complete information on the medications or their side effects.\\
    Score: Doctor 1: 5 points. Doctor 2: 4 points. Doctor 3: 2 points.}} \\
    \hline
    {\makecell[{{>{\raggedright}}p{2cm}}]{Annotator}} & {\makecell[{{>{\raggedright}}p{20cm}}]{Knowledge Correctness: 4 points. Reference: 4 points. Fluency: 4 points.}}\\
    \Xhline{1.0pt}
    \end{tabular}}
    \caption{An example of human annotation results.}
    \label{tab: human annotation example}
    \vspace{-5mm}
\end{table*}

\begin{table*}[htbp]
    \centering
    \resizebox{0.98\textwidth}{!}{
    \begin{tabular}{c|c}
    \Xhline{1.0pt}
    \rowcolor{gray!25} \textbf{Composition} & \textbf{Content} \\
    \hline
    Question & {\makecell[{{>{\raggedright}}p{20cm}}]{What is a cavernous angioma?}} \\
    \hline
    Response ChatDoctor & {\makecell[{{>{\raggedright}}p{20cm}}]{A cavernous angioma is a type of vascular malformation, which are abnormal collections of blood vessels that form in the body and can cause problems such as bleeding, pain, and vision loss. They usually occur in the head or neck but can also be found elsewhere in the body.}} \\
    \hline
    Response Baize Healthcare & {\makecell[{{>{\raggedright}}p{20cm}}]{A cavernous angioma is a type of vascular tumor that forms in the brain. It is also known as a cavernous malformation or a cavernoma.}} \\
    \hline
    Resposne Reference & {\makecell[{{>{\raggedright}}p{20cm}}]{Cavernous angioma appears very similar on magnetic resonance imaging (MRI) with calcified neoplastic and hemorrhagic lesions, most especially renal cell carcinoma, melanoma, pleomorphic xanthoastrocytomas, and oligodendrogliomas.}} \\
    \hline
    Evaluation & {\makecell[{{>{\raggedright}}p{20cm}}]{Analyze:\\Step 1: The patient is asking for information about cavernous angioma.\\Step 2: Doctor 1 provides a general description of cavernous angioma, mentioning that it is a type of vascular malformation and can cause problems. However, it does not specifically address cavernous angioma.\\Step 3: Doctor 2 provides a more accurate description of cavernous angioma, stating that it is a type of vascular tumor that forms in the brain and is also known as a cavernous malformation or a cavernoma.\\Step 4: Doctor 3 provides information on the appearance of cavernous angioma on MRI, but does not provide a clear definition or explanation of what cavernous angioma is.\\Score: Doctor 1: 2 points. Doctor 2: 5 points. Doctor 3: 1 point.}} \\
    \hline
    Reference Evaluation & {\makecell[{{>{\raggedright}}p{20cm}}]{Analyze:\\Step 1: The patient is asking for information about cavernous angioma.\\Step 2: Doctor 1 provides a definition of cavernous angioma, mentioning that it is a benign tumor of the blood vessels, and usually appears in the brain or spine. They also mention the cause and some possible symptoms.\\Step 3: Doctor 2 also provides a definition of cavernous angioma, but focuses more on its occurrence as a type of brain tumor due to abnormal blood vessel growth.\\Step 4: Doctor 3 provides more detailed information about cavernous angioma, including its prevalence, the age range when symptoms usually manifest, and the fact that it can occur de novo, not just congenitally.\\Score: Doctor 1: 5 points. Doctor 2: 3 points. Doctor 3: 5 points.}} \\
    \Xhline{1.0pt}
    \end{tabular}}
    \caption{An incorrect evaluation case with reference evaluation.}
    \label{tab:incorrect evaluation}
\end{table*}

\end{document}